%% file: paper.tex
\documentclass{article}
\usepackage{ijcai25}

\usepackage{times}
\usepackage{soul}
\usepackage{url}
\usepackage[hidelinks]{hyperref}
\usepackage[utf8]{inputenc}
\usepackage[small]{caption}
\usepackage{graphicx}
\usepackage{amsmath}
\usepackage{amsthm}
\usepackage{booktabs}
\usepackage{algorithm}
\usepackage[dvipsnames]{color} % for comments

\usepackage{amsmath,amsfonts,amssymb,amsthm,bm}
\usepackage{mathtools,thmtools,thm-restate}
\usepackage{nicefrac}
\usepackage{array}
\usepackage{comment}
\usepackage{enumitem}
\usepackage[noend]{algpseudocode}
\usepackage[noabbrev,nameinlink,capitalize]{cleveref}
\usepackage[mode=text]{siunitx}
\usepackage{adjustbox,soul}
\usepackage{colortbl}
\usepackage{multirow}
\usepackage{xspace}
\usepackage{subcaption}
\usepackage[dvipsnames]{xcolor}

\usepackage{tikz}
\usepackage{pgf}
\usepackage{forest}

\usetikzlibrary{arrows,decorations.pathmorphing,decorations.footprints,fadings,calc,trees,mindmap,shadows,decorations.text,patterns,positioning,shapes,matrix,fit}
\usetikzlibrary{graphs}
\usetikzlibrary{decorations.pathreplacing}
\usetikzlibrary{arrows,shadows,backgrounds}
\usetikzlibrary{arrows.meta}
\usetikzlibrary{positioning,fit}
\usetikzlibrary{automata}
\usetikzlibrary{matrix}
\usetikzlibrary{karnaugh}
\usetikzlibrary{shapes.symbols,shapes.misc,shapes.arrows}
\usetikzlibrary{matrix,chains,scopes,decorations.pathmorphing}
\usetikzlibrary{fit,calc}

\usepackage[all]{xy}

\input{defs}
\input{preamble}

\begin{document}

\maketitle

\input{abs}

\input{intro}

\input{prelim}
\input{te}

\input{mxp}

\input{res}
\input{conc}
\input{ack}

\input{replbib}
\input{togbbl} % file is automatically generated

% For arxix paper production, and since arXiv does not allow for
% bibtex, we need to create a .bbl file to include upon submission
% to arXiv.
\iftoggle{mkbbl}{
  \bibliographystyle{named}
  \bibliography{team,refs,nfxai}
}{
  % Import bibl (original .bbl) file

\input{paper.bibl}
}

\end{document}

%% file: defs.tex
\input{cdefs}

\newtheorem{proposition}{Proposition}

\newtheorem{example}{Example}
\newtheorem{remark}{Remark}

\crefname{theorem}{Theorem}{Theorems}
\crefname{lemma}{Lemma}{Lemmas}
\crefname{proposition}{Proposition}{Propositions}
\crefname{definition}{Definition}{Definitions}
\crefname{corollary}{Corollary}{Corollaries}
\crefname{example}{Example}{Examples}
\crefname{claim}{Claim}{Claims}
\crefname{assumption}{Assumption}{Assumptions}
\crefname{enumi}{}{}

\newcommand{\mfrk}[1]{{\mathfrak{#1}}}

\newcommand{\fml}[1]{{\mathcal{#1}}}
\newcommand{\set}[1]{\{ #1 \}}

\newcommand{\tbf}[1]{\textbf{#1}}
\newcommand{\mbf}[1]{\ensuremath\mathbf{#1}}
\newcommand{\msf}[1]{\ensuremath\mathsf{#1}}
\newcommand{\mbb}[1]{\ensuremath\mathbb{#1}}

\newcommand{\iaxp}{iAXp\xspace}
\newcommand{\icxp}{iCXp\xspace}
\newcommand{\mxiaxp}{Max-iAXp\xspace}

\newcommand{\bigland}{\ensuremath\bigwedge}

\newcounter{tableeqn}[table]

\DeclareMathOperator*{\limply}{\rightarrow}

\DeclareMathOperator*{\argmax}{\msf{argmax}}

\newcommand{\jnoteF}[1]{}

\newcolumntype{L}[1]{>{\raggedright\let\newline\\\arraybackslash\hspace{0pt}}m{#1}}
\newcolumntype{C}[1]{>{\centering\let\newline\\\arraybackslash\hspace{0pt}}m{#1}}
\newcolumntype{R}[1]{>{\raggedleft\let\newline\\\arraybackslash\hspace{0pt}}m{#1}}

\tikzset{
  0 my edge/.style={densely dashed, my edge},
  my edge/.style={-{Stealth[]}},
}

\def\HiLi{\leavevmode\rlap{\hbox to \linewidth{\color{platinum}\leaders\hrule height .8\baselineskip depth .5ex\hfill}}}

\setitemize{noitemsep,topsep=0pt,parsep=0pt,partopsep=0pt}
\setenumerate{noitemsep,topsep=0pt,parsep=0pt,partopsep=0pt}
\setlist{nosep,leftmargin=0.45cm}

\providetoggle{long}
\settoggle{long}{false}

\makeatletter
\algnewcommand{\LineComment}[1]{\Statex \hskip\ALG@thistlm \(\triangleright\) #1}
\makeatother

\newcommand{\True}{\textbf{true}}
\newcommand{\False}{\textbf{false}}

\newcommand{\MxSAT}{\ensuremath\mathsf{MaxSAT}}
\newcommand{\outc}{\ensuremath\mathsf{hasCEx}} %\mathsf{hasAEx}

\newcommand{\ignore}[1]{}

\newcommand{\bb}[1]{\ensuremath{[\![#1]\!]}}

\definecolor{midblue}{rgb}{0.2,0.2,0.7}
\definecolor{platinum}{rgb}{0.9, 0.89, 0.89} %#E5E4E2
\definecolor{midgrey}{rgb}{0.5,0.5,0.5}

%% file: cdefs.tex
\definecolor{gray}{rgb}{.4,.4,.4}
\definecolor{midgrey}{rgb}{0.5,0.5,0.5}
\definecolor{middarkgrey}{rgb}{0.35,0.35,0.35}
\definecolor{darkgrey}{rgb}{0.3,0.3,0.3}
\definecolor{darkred}{rgb}{0.7,0.1,0.1}
\definecolor{midblue}{rgb}{0.2,0.2,0.7}
\definecolor{darkblue}{rgb}{0.1,0.1,0.5}
\definecolor{defseagreen}{cmyk}{0.69,0,0.50,0}

\definecolor{tyellow1}{HTML}{FCE94F}
\definecolor{tyellow2}{HTML}{EDD400}
\definecolor{tyellow3}{HTML}{C4A000}
\definecolor{torange1}{HTML}{FCAF3E}
\definecolor{torange2}{HTML}{F57900}
\definecolor{torange3}{HTML}{C35C00}
\definecolor{tbrown1}{HTML}{E9B96E}
\definecolor{tbrown2}{HTML}{C17D11}
\definecolor{tbrown3}{HTML}{8F5902}
\definecolor{tgreen1}{HTML}{8AE234}
\definecolor{tgreen2}{HTML}{73D216}
\definecolor{tgreen3}{HTML}{4E9A06}
\definecolor{tblue1}{HTML}{729FCF}
\definecolor{tblue2}{HTML}{3465A4}
\definecolor{tblue3}{HTML}{204A87}
\definecolor{tpurple1}{HTML}{AD7FA8}
\definecolor{tpurple2}{HTML}{75507B}
\definecolor{tpurple3}{HTML}{5C3566}
\definecolor{tred1}{HTML}{EF2929}
\definecolor{tred2}{HTML}{CC0000}
\definecolor{tred3}{HTML}{A40000}
\definecolor{tlgray1}{HTML}{EEEEEC}
\definecolor{tlgray2}{HTML}{D3D7CF}
\definecolor{tlgray3}{HTML}{BABDB6}
\definecolor{tdgray1}{HTML}{888A85}
\definecolor{tdgray2}{HTML}{555753}
\definecolor{tdgray3}{HTML}{2E3436}

%% file: preamble.tex
\title{Most General Explanations of Tree Ensembles (Extended Version)}

\author{
    Yacine Izza\textsuperscript{\rm 1}\thanks{Corresponding Author. Email: izza@comp.nus.edu.sg} \and
    Alexey Ignatiev\textsuperscript{\rm 2} \and
    Sasha Rubin\textsuperscript{\rm 3} \and 
    Joao Marques-Silva\textsuperscript{\rm 4} \and
    Peter J. Stuckey\textsuperscript{\rm 2,5}\\
\affiliations
    \textsuperscript{\rm 1}CREATE, NUS, Singapore \and
    \textsuperscript{\rm 2}Monash University,  Australia\\
    \textsuperscript{\rm 3} The University of Sydney, Australia \and
    \textsuperscript{\rm 4} ICREA, University of Lleida, Spain\\
    \textsuperscript{\rm 5}OPTIMA ARC Industrial Training and Transformation
    Centre, Australia\\  
}

%% file: abs.tex
%------------------------------------------------------------------------------%
%File:        abs.tex
%------------------------------------------------------------------------------%
%
\begin{abstract}
  %
  %Explainable Artificial Intelligence (XAI) has arisen as an important
  %field to answer concerns about decisions made by black-box AI
  %systems.
  Explainable Artificial Intelligence (XAI) is critical for
  attaining trust in the operation of AI systems.
  A key question of an AI system is ``why was this decision
  made this way''.
  Formal approaches to XAI use a formal model of the AI system
  to identify \emph{abductive explanations}.
  %to reason about its properties.
  %
  %A formal \emph{abductive explanation} for a decision by an AI system
  %on an instance is defined as a minimal set of inputs to the system
  %that, if they take the same concrete values as the instance being
  %explained, will ensure the AI system reaches the same decision.
  %
  While abductive explanations %are global explanations potentially
  may be applicable to a large number of inputs sharing the same
  concrete values, more general explanations may be preferred for
  numeric inputs.
  So-called \emph{inflated abductive explanations} give intervals for
  each feature ensuring that any input whose values fall withing these
  intervals is still guaranteed to make the same prediction.
  %        \sr{the
  %          difference between abductive explanations and inflated abductive
  %          explanations is not so clear from the abstract.}\pjs{Improved'?}
  %
  % for an instance, is a subset of values for each input that
  % includes the input value for that instance, which guarantees that
  % the same decision is made for all choices of inputs which lie
  % within these subsets.
  %
  Inflated explanations cover a larger portion of the input space, and
  hence are deemed more general explanations.
  But there can be many (inflated) abductive explanations for an
  instance.
  Which is the best?
  In this paper, we show how to find a most general abductive
  explanation for an AI decision.
  This explanation covers as much of the input space as possible,
  while still being a correct formal explanation of the model's
  behaviour.
  Given that we only want to give a human one explanation for a decision,
  the most general explanation gives us the explanation with the
  broadest applicability, and hence the one most likely to seem
  sensible.
\end{abstract}
%
%------------------------------------------------------------------------------%

%% file: intro.tex
\section{Introduction} \label{sec:intro}

The widespread use of artificial intelligence to make or support decisions
effecting humans has led to the need for these systems to be able to
explain their decisions. The field of eXplainable AI (XAI) has emerged in
response to this need. Its generally accepted that XAI is required to
deliver trustworthy AI systems.
Unfortunately the bulk of work in XAI offers no formal guarantees, or even
clear definitions of the ``meaning'' of an explanation. 
Examples of non-formal XAI include model-agnostic
methods~\cite{guestrin-kdd16,lundberg-nips17,guestrin-aaai18},
heuristic learning of saliency maps (and their
variants)~\cite{muller-plosone15,muller-xai19-ch01,xai-bk19,muller-ieee-proc21},
but also proposals of intrinsic
interpretability~\cite{rudin-naturemi19,molnar-bk20,rudin-ss22}.
In recent years, comprehensive evidence has been gathered that attests
to  the lack of rigor of these (non-formal) XAI
approaches~\cite{Lakkaraju-aies20,ignatiev-ijcai20,barrett-nips23,msh-cacm24,hms-jar24}. 
%\cite{ignatiev-ijcai20,Lakkaraju-aies20,ims-ijcai21,iims-jair22,barrett-nips23,msh-cacm24}

Formal XAI~\cite{darwiche-ijcai18,inms-aaai19,msi-aaai22} 
in contrast provides rigorous definitions of explanations which
have desirable properties as well as practical algorithms to compute them.
But formal XAI methods also have limitations: they may not scale and
hence may be unable to provide explanations for some complex AI models. 

Abductive explanations~\cite{darwiche-ijcai18,inms-aaai19} are the most
commonly used formal XAI approach. An abductive explanation is a
subset-minimal set of features, which guarantee that if the values of the
instance being explained are used for these features, then the decision of
the AI system will remain the same. Hence these feature values of the
instance are ``sufficient'' to ensure the decision.
For example if a male patient with O+ blood type, height 1.82m and weight
90kg is found to be at risk of a disease by an AI system, an abductive
explanation might be that the height and weight are the only required
features:
thus anyone with height 1.82m and weight 90kg would also be at risk.

While abductive explanations have nice properties and are a valuable tool,
they are quite restrictive. The abductive explanation above says nothing
about a person with height 1.81m.  
Recently, \emph{inflated formal explanations} have been
proposed~\cite{iisms-aaai24},  %\cite{iisms-corr23a,iisms-aaai24}
which extend abductive explanations to include maximal ranges
for features that still guarantee the same decision.
An inflated abductive explanation for the disease example may be that
any person with height in the range 1.80-2.50m and weight 90-150kg also has
the same risk. 

When explaining the reason for an AI system decision for a particular
instance, there can exist many possible formal abductive explanations; 
similarly there can exist many possible inflated abductive explanations.
Ideally when explaining a decision to a human we would like to give a single
explanation.  In this work we show how we can compute the \emph{most general
abductive explanation}, in some sense the most general explanation of the AI
systems behaviour.
More precisely, the contributions of this paper are:

\begin{enumerate}
    \item An %elegant 
    	implicit hitting set approach to compute most general inflated abductive explanations;
    \item Two Boolean Satisfiability (SAT) encodings and an Integer Linear
      Program (ILP) encoding for generating candidate maximal abductive explanations; 
    \item Experiments showing that we can in practice create most general 
    abductive explanations; 
    \item A unified propositional encoding for tree ensembles. 
\end{enumerate}

This CoRR report is an extended version of our IJCAI'25 paper, providing a detailed 
presentation of our general propositional encoding applicable to any 
tree ensemble model, including random forests with weighted votes (i.e., probability scores).
We emphasize that the source code of our MaxSAT-based approach 
to explaining  random forests with weighted votes has been publicly 
available in the GitHub repository \texttt{RFxpl}  since 
Nov'2022  --- verifiable via the repository history.\footnote{Together with the CoRR 
report, \texttt{RFxpl} repository serves as a timestamp to safeguard against potential 
 scooping of our contribution.}

%% file: prelim.tex
\section{Preliminaries} \label{sec:prelim}

\subsection{Classification Problems}
Let $\fml{F}$ be a set of variables called \emph{features}, say $\fml{F}=[m]$.
Each feature $i$ is equipped with a finite \emph{domain} $\mbb{D}_i$.
In general, when referring to the value of a feature $i\in\fml{F}$, we
will use a variable $x_i$, with $x_i$ taking values from $\mbb{D}_i$.
%
%Domains can be categorical or ordinal, with
%values that can be boolean, integer or real-valued.
Domains are ordinal that can be integer or real-valued.
The {\it feature space} is defined as $\mbb{F}=\prod\nolimits_{i\in\fml{F}} \mbb{D}_i$.
The notation $\mbf{x}=(x_1,\ldots,x_m)$ denotes an arbitrary point in
feature space, whilst the notation $\mbf{v}=(v_1,\ldots,v_m)$
represents a specific point in feature space.
A \emph{region} is a set $\mbb{E} = \mbb{E}_1 \times \cdots \times \mbb{E}_m$
consisting of, for each feature $i$, a non-empty set
$\mbb{E}_i \subseteq \mbb{D}_i$ of values
(it can also be viewed as a function that maps feature $i$ to $\mbb{E}_i$).
A \emph{total classification function}
$\kappa:\mbb{F}\to\fml{K}$ where $\fml{K}=\{c_1,\ldots,c_K\}$ is
a set of \emph{classes} for some $K \geq 2$.
For technical reasons, we also require $\kappa$ not to be a constant
function, i.e.\ there exists at least two points in feature space with
differing predictions.
An \emph{instance} denotes a pair $(\mbf{v}, c)$, where
$\mbf{v}\in\mbb{F}$ and $c\in\fml{K}$.
We represent a classifier by a tuple
$\fml{M}=(\fml{F},\mbb{F},\fml{K},\kappa)$.
Given the above, an \emph{explanation problem} is a tuple
$\fml{E} = (\fml{M},(\mbf{v},c))$.%, where $\kappa(\mbf{v}) = c$.

\subsection{Formal Explainability}
Two types of formal explanations have been predominantly studied:
abductive~\cite{darwiche-ijcai18,inms-aaai19} and
contrastive~\cite{miller-aij19,inams-aiia20}.
Abductive explanations (AXps) broadly answer a \tbf{Why} question,
i.e.\ \emph{Why the prediction?}, whereas contrastive explanations
(CXps) broadly answer a \tbf{Why Not} question, i.e.\ \emph{Why not
some other prediction?}.
Intuitively, an AXp is a subset-minimal set of feature values
$(x_i=v_i)$, at most one for each feature $i\in\fml{F}$, that is
sufficient to trigger a particular class and satisfy the instance
being explained.
Similarly, a CXp is a subset-minimal set of features by changing the
values of which one can trigger a class different from the target one.
More formally, AXps and CXps are defined below.

Given an explanation problem $\fml{E}=(\fml{M}, (\mbf{v}, c))$, an
\emph{abductive explanation (AXp) of $\fml{E}$} is a subset-minimal set
$\fml{X}\subseteq\fml{F}$ of features which, if assigned the values dictated by
the instance $(\mbf{v},c)$, are sufficient for the prediction.
The latter condition is formally stated, for a set $\fml{X}$, as follows:
\begin{equation} \label{eq:axp}
\forall(\mbf{x}\in\mbb{F}).\left[\bigland\nolimits_{i\in\fml{X}}(x_i=v_i)\right]\limply(\kappa(\mbf{x})=c)
\end{equation}
Any subset $\fml{X} \subseteq \fml{F}$ that satisfies \eqref{eq:axp},
but is not subset-minimal (i.e.\ there exists $\fml{X}^\prime \subset
\fml{X}$ that satisfies~\eqref{eq:axp}), is referred to as a
\emph{weak abductive explanation (Weak AXp)}.

Given an explanation problem, a contrastive explanation
(CXp) is a subset-minimal set of features $\fml{Y}\subseteq\fml{F}$
which, if the features in $\fml{F}\setminus\fml{Y}$ are assigned the
values dictated by the instance $(\mbf{v},c)$, then there is an
assignment to the features in $\fml{Y}$ that changes the prediction.
This is stated as follows, for a chosen set $\fml{Y}\subseteq\fml{F}$:
\begin{equation} \label{eq:cxp}
  \exists(\mbf{x}\in\mbb{F}).\left[\bigland\nolimits_{i\in\fml{F}\setminus\fml{Y}}(x_i=v_i)\right]\land(\kappa(\mbf{x})\not=c)
\end{equation}

AXp's and CXp's respect a minimal-hitting set (MHS) duality
relationship~\cite{inams-aiia20}.
Concretely, each AXp is an MHS of the CXp's and each CXp is an MHS of
the AXp's.
MHS duality is a stepping stone for the enumeration of explanations.

\paragraph{Inflated Formal Explanations.}
An \emph{inflated abductive explanation} (\iaxp) of 
$\fml{E} = (\fml{M},(\mbf{v},c))$ is a tuple 
$(\fml{X},\mbb{E})$, where
$\fml{X}\subseteq\fml{F}$ is an AXp, and $\mbb{E}$ is a region
where (i) $\mbb{E}_i \subset \mbb{D}_i$ %$\mbb{E}_i \subsetneq \mbb{D}_i$
for each $ i\in\fml{X}$ and  $\mbb{E}_i = \mbb{D}_i$ for each
$i\in(\fml{F}\setminus\fml{X})$, (ii) $v_i \in \mbb{E}_i$ for each
$i \in \fml{X}$, such that
  \begin{equation} \label{eq:iaxp}
    \forall(\mbf{x}\in\mbb{F}).
    \left[
      \bigland\nolimits_{i\in{\fml{X}}}(x_i\in \mbb{E}_i)
      \right]
    \limply(\kappa(\mbf{x})=c)
  \end{equation}
and $\mbb{E}$ is a {\it maximal set} with properties (i), (ii),
and~\eqref{eq:iaxp}, i.e.\  if $\mbb{E}'$ is any range that satisfies (i), (ii)
and \eqref{eq:iaxp}, then it is not the case that $\mbb{E}' \supset \mbb{E}$.

A \emph{size measure} $s$ is a mononotic function that maps a region
$\mbb{E}\subseteq\mbb{F}$ to a real number.
Consider an \iaxp $(\fml{X}, \mbb{E})$ and a size
measure $s$, then $(\fml{X}, \mbb{E})$ is called a {\it maximum iAXp}
if its size score $s(\mbb{E})$ is a maximum amongst all \iaxp's, i.e.,
if $\mbb{E}'$ is an \iaxp, then $s(\mbb{E}') \leq s(\mbb{E})$.

Later, when dealing with tree ensemble models, we will instantiate $s$
to a specific coverage measure on ranges consisting of intervals.

\begin{example}
  \label{ex:iaxp}
  Consider 3 features $x_i$, $i\in[3]$, representing an
  individual's blood type, their age, and weight with the domains
  $\mbb{D}_1=\{A, B, AB, O\}$, $\mbb{D}_2=[20, 80]$, and
  $\mbb{D}_3=[50, 150]$.
  Assume the classifier $\kappa(\mbf{x})$ predicts a high risk of a
  disease if $x_2 \geq 60$ and $x_3 \geq 80$;
  otherwise, $\kappa(\mbf{x})$ predicts the individual to be at low
  risk of a disease.
  (Observe that the blood type $x_1$ is ignored in the decision making
  process.)
  Consider an instance $\mbf{v}=(A, 65, 85)$, which is classified as
  high risk.
  The only AXp for this instance is $\fml{X}=\{2,3\}$.
  Now, consider a simple size measure
  $s(\mbb{E}_i)=\nicefrac{|\mbb{E}_i|}{|\mbb{D}_i|}$.
  While there are multiple ways to inflate $\fml{X}$ using this size
  measure, the \emph{maximum} and, intuitively, the most general iAXp
  consists of the intervals $\mbb{E}_2=[60, 80]$ and
  $\mbb{E}_3=[80, 150]$.
\end{example}

%Conversely to inflated AXp, 
An {\it inflated contrastive explanation} (\icxp) is a
pair $(\fml{Y}, \mbb{G})$ s.t.\ $\fml{Y}$ is a CXp of $\fml{E}$,  and
$v_i \not\in \mbb{G}_i$ for each feature $i\in\fml{Y}$ and $\mbb{G}_i
= \set{v_i}$ for any $i\in(\fml{F}\setminus\fml{Y})$, such that the
following holds:
\begin{align} \label{eq:icxp}
  \exists(\mbf{x}\in\mbb{F}).
  \left[\bigland\nolimits_{i\in\fml{F}}(x_i \in \mbb{G}_i) \right]%
  \land(\kappa(\mbf{x})\not=c) %\nonumber
\end{align}

\begin{example}
  \label{ex:icxp}
  Consider again \cref{ex:iaxp} and $\mbf{v}=(A, 65, 85)$ classified 
  as a high risk.
%  Getting back to Example~\ref{ex:iaxp}, recall that instance
%  $\mbf{v}=(A, 65, 85)$ is classified as high risk of a disease.
  %
  If a user is interested in answering \emph{why not} a low risk, a
  possible CXp to extract is $\fml{Y}_1=\{2\}$.
  This means that by suitably changing the value of $x_2$ to some value 
  taken from its entire domain $[20, 80]$, the prediction can
  be changed to low risk even if features $x_1$ and $x_3$ are fixed, resp., to
  values $A$ and $85$.
  Observe that the freedom of selecting the entire domain $\mbb{D}_2$
  although perfectly valid provides a user with little to no insight
  on how the prediction can be changed, as it contains values, namely
  those in $[60, 80]$, that if chosen do not lead to a
  misclassification.
  Inflating the CXp $\fml{Y}_1$ can be done differently and may result in a
  valid shrunk interval, say, $\mbb{G}_2=[20, 65]$.
  Intuitively, the most informative \icxp shrinks the
  domain $\mbb{D}_2$ into the set $\mbb{G}_2^{'}=[20,60)$, while
  ensuring a misclassification is still achievable.
  Similarly, another CXp for instance $\mbf{v}$ is $\fml{Y}_2=\{3\}$,
  with the most sensible inflation being $\mbb{G}_3=[50, 80)$.
\end{example}

Let us denote the set of all \iaxp's for a particular explanation
problem $\fml{E}$ as $\mbb{A}(\fml{E})$ while the set of all \icxp's
will be denoted as $\mbb{C}(\fml{E})$.
Similar to the case of traditional abductive and contrastive
explanations, minimal hitting set (MHS) duality between \iaxp’s and
\icxp’s was established in earlier work \cite{iisms-aaai24}.
%\cite{iisms-corr23a,iisms-aaai24}.

\begin{proposition} \label{prop:inflated-duality}
  Given an explanation problem $\fml{E}$, each \iaxp
  $(\fml{X},\mbb{E})\in\mbb{A}(\fml{E})$ minimally ``hits'' each \icxp
  $(\fml{Y},\mbb{G})\in\mbb{C}(\fml{E})$ s.t. if feature $i\in\fml{F}$
  is selected to ``hit'' \iaxp $(\fml{X},\mbb{E})$ then
  $\mbb{G}_i\cap\mbb{E}_i=\emptyset$, and vice versa.
\end{proposition}

\begin{remark}
  As the previous two examples demonstrate, inflation of an AXp or a
  CXp can be done in multiple ways resulting in different inflated
  AXps or CXps, respectively.
  While many of those inflated explanations are valid, some may have a
  high degree of redundancy, which is often undesirable in practice.
  Therefore, the goal of explanation inflation is (i)~to expand the
  intervals included in an \iaxp as much as possible and (ii)~to shrink
  the intervals included in an \icxp as much as possible, providing the
  most general way of (i) ensuring the prediction and (ii) breaking
  it, respectively.
  The above proposition establishes a duality between inflated AXps
  and CXps assuming they are inflated in \emph{the most sensible} way,
  with no redundancy involved, hence the requirement $\mbb{G}_i\cap
  \mbb{E}_i=\emptyset$.
\end{remark}

\autoref{prop:inflated-duality} forms the foundation of the algorithm
proposed in this work for computing a maximum size inflated abductive
explanations  
inspired by the earlier work in the area of implicit hitting set
enumeration~\cite{bacchus-cp11,iplms-cp15,jarvisalo-kr16}.
%\cite{bacchus-cp11,iplms-cp15,jarvisalo-kr16,imms-ecai16}. %refs TO be included in FINAL version

\subsection{SAT and MaxSAT}
We assume standard definitions for propositional satisfiability (SAT)
and maximum satisfiability (MaxSAT) solving~\cite{sat-handbook21}.
The \emph{maximum satisfiability (MaxSAT) problem} is to find
a truth assignment that maximizes the number of satisfied
propositional formulas in a clausal form (i.e. CNF formula).
There are a number of variants of
MaxSAT~\cite[Chapters~23~and~24]{sat-handbook21}.
We will be mostly interested in \emph{Partial Weighted MaxSAT},
which can be formulated as follows.
The input WCNF formula $\phi$ is of the form $\langle\fml{H},\fml{S}\rangle$
%$\fml{H}\land\fml{S}$
where $\fml{H}$ is a set of \emph{hard} clauses, which
must be satisfied, and $\fml{S}$ is a set of \emph{soft} clauses,
each with a weight, which represent a preference to satisfy
those clauses.
Whenever convenient, a soft clause $c$ with weight $w$ will be denoted
by $(c, w)$.
The \emph{Partial Weighted MaxSAT problem} consists in finding an assignment
that satisfies all the hard clauses and maximizes the sum of the weights of
the satisfied soft clauses.
%

%% related work

\input{relw}

%% file: relw.tex
%
\subsection{Related Work}

Logic-based explanations of ML models have been studied since
2018~\cite{darwiche-ijcai18}, with recent surveys summarizing the
observed progress~\cite{msi-aaai22,ms-rw22,darwiche-lics23,ms-isola24}.
However, the study of logic-explanations in AI can at least be traced
to the late 1970s and
1980s~\cite{swartout-ijcai77,swartout-aij83,shanahan-ijcai89}.

%The complexity of computing subset- and cardinality-minimal
%explanations have been the studied in several
%works~\cite{ims-ijcai21,ims-sat21,cms-aij23,katz-icml24}.

%Inflated explanations were proposed by \cite{iisms-aaai24,iisms-corr23a}. 
%%2023~\cite{iisms-aaai24,iisms-corr23a}
%Nevertheless, the use of more general literals in abductive
%explanations was studied in other
%works~\cite{iims-jair22,darwiche-jelia23}. 

Inflated explanations are more general explanations than AXps. This
has been recognized in the earlier work in model-agnostic
explainability (Anchor~\cite{guestrin-aaai18}) 
and in the formal explainability~\cite{iims-jair22,darwiche-jelia23,YuIS23,iisms-corr23a,iisms-aaai24}.

Implicit hitting-set algorithms have been used in a wide range of
practical
domains~\cite{stuckey-padl05,karp-soda11,bacchus-cp11,pms-aaai13,lm-cpaior13,iplms-cp15,lpmms-cj16,jarvisalo-kr16}.
In the case of XAI, implicit hitting-set algorithms, mimicking
MARCO~\cite{lpmms-cj16}, have been applied in the enumeration of
abductive explanations~\cite{ms-rw22} 
but also for deciding feature relevancy~\cite{hims-aaai23}.
Implicit hitting-set algorithms can be viewed as a variant of the
general paradigm of counterexample-guided abstraction refinement
(CEGAR), which was originally devised in the context of
model-checking~\cite{clarke-jacm03}.

%Logic-based explainability is covered in a number of recent works.
%
Additional recent works on formal explainability for large scale ML 
models include~\cite{barrett-nips23,katz-tacas23,swiftxp-kr24,barrett-corr24,ims-corr24c}.
Probabilistic explanations are investigated
in~\cite{kutyniok-jair21,ihincms-ijar23,izza-ecai24}.
There exist proposals to account for constraints on the
inputs~\cite{rubin-aaai22,yisnms-aaai23}. %%
An extension of feature selection-based formal explanations 
into feature attribution-based formal explanations 
is proposed in~\cite{ignatiev-corr23a,izza-aaai24b,Ignatiev-sat24b,lhms-aaai25}.

%% file: te.tex
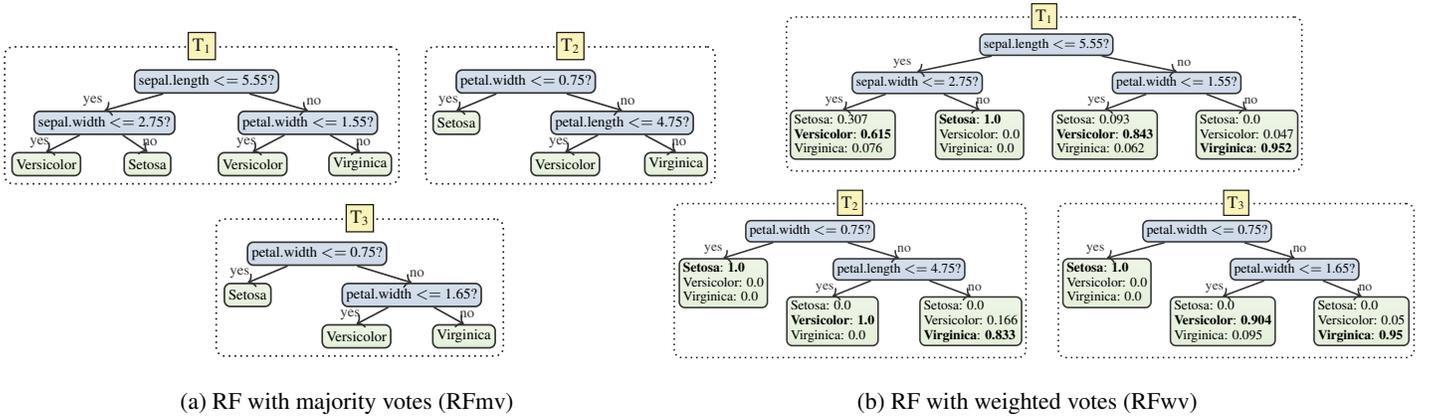
\begin{figure*}[ht]
  \begin{subfigure}{0.43\textwidth}
    \centering
    \scalebox{0.79}{\input{./texfigs/ex3.tex}}
    \caption{RF with majority votes (RFmv)}
    \label{fig:RFmv}
  \end{subfigure}
  \begin{subfigure}{0.58\textwidth}
    \centering
    \scalebox{0.73}{\input{./texfigs/ex2.tex}}
    \caption{RF with weighted votes (RFwv)}
    \label{fig:RFwv}
  \end{subfigure}
\caption{
    Example of a simple RF trained using Scikit-learn with 3 trees, on {\it Iris} dataset. 
    %Here, the tree ensemble has 3 trees  with the depth of each tree being at most 2.
%    On left is reported an RF with majority vote prediction, and on right 
%    an RF with weighted votes. 
    On left is reported an RFmv classifier and on right 
    an RFwv.     
    In the running examples, classes ``Setosa'', ``Versicolor'' and  ``Virginica'' are 
    denoted, resp.\  class $c_1$, $c_2$ and $c_3$, and features {\it sepal.length},  
    {\it sepal.width}, {\it petal.length} and  {\it petal.width} are ordered, resp., as 
    features 1--4.
} \label{fig:RF}
\end{figure*}

\section{Tree Ensembles}
This section proposes a unified representation for tree ensembles
(TEs). The unified representation will be instrumental when relating
concrete instantiations of TEs.

\paragraph{General TE model.}
We propose a general model for selecting a class in a tree ensemble
(TE). As shown below, the proposed model serves to represent boosted
trees~\cite{friedman-tas01}, random forests with majority
voting~\cite{Breiman01}, and random forests with weighted
voting (Scikit-learn~\cite{scikitlearn}).

A decision tree $T$ on features $\fml{F}$ for classes $\fml{K}$
is a binary-branching rooted-tree whose internal nodes are labeled 
by \emph{split conditions}   
which are predicates of the form $x_i < d$ for some feature 
$i \in \fml{F}$ and $d \in \mbb{D}_i$, and leaf nodes labelled by a class
$c \in \fml{K}$. 
A path $R_\mfrk{n}$ to a leaf node $\mfrk{n}$ can be considered
as the set of split conditions 
%(when going left-condition holds) or their
%negations (when going right-condition fails) of the split nodes 
of nodes 
visited on the path from root to $\mfrk{n}$.

A tree ensemble $\mfrk{T}$ has $n$ decision trees, $T_1,\ldots,T_n$. 
Each path
$R_{l}$ in each decision tree $T_t$ is assigned a class $cl(l)$
(corresponding to the class of the leaf node  it reaches) and a
weight $w_{l}$.
Given a point $\mbf{v}$ in feature space, and for each decision tree
$T_t$, with $1\le{t}\le{n}$, there is a single path $R_{k_{\mbf{v},t}}$
that is consistent with $\mbf{v}$ in tree $T_t$.
The index $k_{\mbf{v},t}$
denotes which path of $T_t$ is consistent with $\mbf{v}$.

For decision tree $T_t$, and given $\mbf{v}$, the picked class is 
denoted $cl(k_{\mbf{v},t})$  and the picked weight $w_{k_{\mbf{v},t}}$.
For each class $c \in\fml{K}$, the class weight will be computed by,
\[
W(c,\mbf{v})=\sum\nolimits_{t=1}^{n} \textbf{if}~ c=cl(k_{{\mbf{v}},t})
\textbf{~then~} w_{k_{{\mbf{v}},t}} \textbf{~else~} 0
\]
Finally, the picked class is given by,
\[
\kappa(\mbf{v}) = \argmax\nolimits_{c\in\fml{K}}\{W(c,\mbf{v})\} %\,|\,c_r\in\fml{K}
\]

\paragraph{Random forests with majority voting.}
In a random forest (RF) with majority voting (RFmv),  %(see~\cite{Breiman01}),
and given a point in feature space, each decision tree picks a class;
the final picked class is the one picked by most trees.
Note that to represent an RFmv in our general TE model, we set 
the weights of the leaf nodes to 1.

%To represent an RFmv in our general TE model, we
%proceed as follows. The weight of every path in each tree is 1.
%Moreover, the class output by each tree path in the TE is the class
%predicted by the corresponding tree path in the RF.

\begin{example} \label{ex:RF(mv)}
  \cref{fig:RFmv} shows an example of a RFmv of 3 trees, trained 
  with Scikit-learn on {\it Iris} dataset. 
  Consider a sample $\mbf{v} = (6.0, 3.5, 1.4, 0.2)$ 
  then the counts of votes for classes 1--3 are, resp.,
  $W(c_1,\mbf{v}) = 2$ (votes of T$_2$ and T$_3$), $W(c_2,\mbf{v}) =1$ 
  (vote of T$_1$),   and  $W(c_3,\mbf{v}) = 0$. 
  Hence, the predicted class is $c_1$ (``Setosa'') as it has the highest
  score.
\end{example}

%Hence, the class counting a score greater or equal 2 is the wining one.

\paragraph{Random forests with weighted voting.}
%~\\
%
In an RF with weighted voting (RFwv), and given a point in feature space,
each tree picks a weight for a chosen selected class; the final picked
class is the class with the largest weight.

\begin{example} \label{ex:RF(wv)}
  \cref{fig:RFwv} shows an example of a RF with weighted voting, which 
  has 3 trees.
  Consider the sample $\mbf{v} = (5.1, 3.5, 1.4, 0.2)$  
  then the scores of classes 1--3 are $W(c_1,\mbf{v}) = 1.0+1.0+1.0 = 3.0$,
  $W(c_2,\mbf{v}) =  0.0$, and  $W(c_3,\mbf{v}) =  0.0$. 
  Hence, the predicted class is $c_1$ (``Setosa'') as it has the highest score.
\end{example}

%To represent an RFwv, in our general TE model, we
%proceed as follows. The weight of each tree path in the TE model is
%the one in the corresponding tree path in the RF. Moreover, the class
%output by the path in the TE is the class predicted by the
%corresponding tree path in the RF.

\paragraph{Boosted trees.}
In a boosted tree~(BT, see~\cite{friedman-tas01}), and given a point
in feature space, class selection works as follows: each tree is
pre-assigned a class $c$ (all leaves have the same class $c$) 
and returns a weight; the weights of each class
are summed up over all trees, and the class with the largest weight is
picked.
%

%To represent a BT in our general TE model, we proceed as follows.
%Any tree in the BT will be mapped to a tree in the TE model.
%Each tree path weight in the BT will be mapped to the same tree
%path weight in the TE model. Finally, for each decision tree in the
%group of trees assigned to class $c$ in the BT, any path in the
%corresponding tree in the TE model will output class $c$.

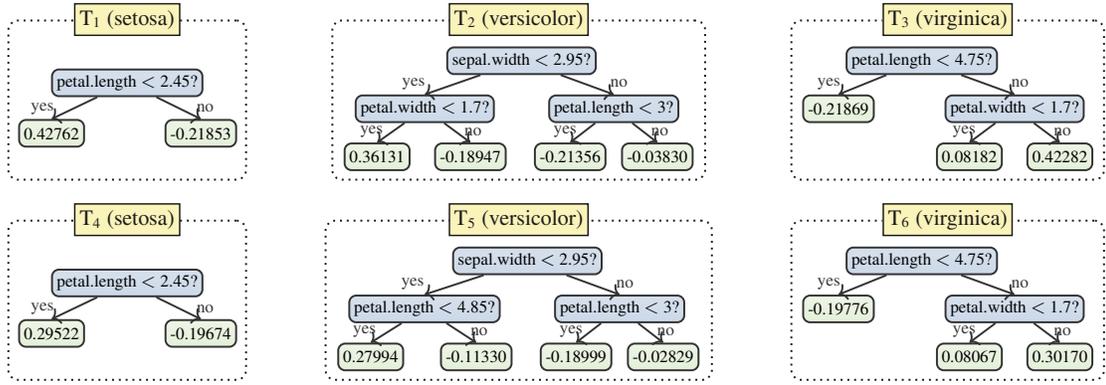
\begin{figure*}[ht]
  \begin{center}
    \resizebox{0.95\textwidth}{!}{\input{./texfigs/ex1.tex}}
    \caption{ Example of a simple boosted tree
      (BT) generated by XGBoost on Iris dataset, s.t. 
      number of trees per class is set to 2.
      %ensemble has 2 trees for each of the 3 classes with the depth of
      %each tree being at most 2.
      }
    \label{fig:BT}
  \end{center}
\end{figure*}

\begin{example} \label{ex:bt}
  An example of a BT trained by XGBoost for the Iris dataset is shown
  in Figure~\ref{fig:BT}.
  %
%  The dataset includes 4 numeric features:  \emph{sepal.length},  \emph{sepal.width},
%  \emph{petal.length},  \emph{petal.width}, and
%  3 classes: $\text{\em class}_\text{\em 1}=\text{\em``setosa''}$,
%  $\text{\em class}_\text{\em 2}=\text{\em``versicolor''}$, and
%  $\text{\em class}_\text{\em 3}=\text{\em``virginica''}$.
  %
  Each class $i\in[3]$ is represented in the BT by 2 trees $T_{3j+i}$,
  $j\in\{0, 1\}$.
  This way, given an input instance $(\text{\em
  sepal.length}=\text{\em 5.1}) \land (\text{\em
  sepal.width}=\text{\em 3.5}) \land (\text{\em petal.length}=\text{\em
  1.4}) \land (\text{\em petal.width}=\text{\em 0.2})$, the scores of
  classes 1--3 are $w_1=\fml{T}_1+\fml{T}_4=0.72284$,
  $w_2=\fml{T}_2+\fml{T}_5=-0.40355$, and
  $w_3=\fml{T}_3+\fml{T}_6=-0.41645$.
  Hence, the model predicts class~1 (``setosa'') as it has the highest
  score.
  The path taken by this instance in tree $T_2$ is defined as $\{
  \text{\em sepal.width} \geq \text{\em 2.95}, \text{\em petal.length} <
  \text{\em 3} \}$.
\end{example}

\begin{remark}
   It is simple to reduce RFmv to RFwv, but also to BTs.\\
  The reduction from RFmv to RFwv is straightforward, in that the
  weights are just set to 1. \\
  The reduction from RFmv to BT also seems simple. A tree $T$ in the
  RF is split into $L$ trees, one for each class predicted by $T$.
  For copy $l$, added to the group representing class $c_l$, the paths
  predicting $c_l$ are given weight 1; all other paths are given
  weight 0.
\end{remark}

\begin{proposition}
  For each random forest with majority voting there exists an
  equivalent boosted tree with a size that is polynomial on the size
  of the starting random forest.
\end{proposition}

\begin{proposition}
  For each random forest with majority voting there exists an
  equivalent random forest with weighted voting with a size that is
  linear on the size of the starting random forest.
\end{proposition}

%%%%%%%%%%%%%%%%%%%%%%%%%%%%%%%%
\input{tenc}

%% file: texfigs/ex3.tex
%------------------------------------------------------------------------------%
% File:        beacon.tex
%
% Description:
%
% Created:     7 Feb 2018.
%
% Author:      Alexey Ignatiev (aign).
%------------------------------------------------------------------------------%

% defining box structure
\tikzstyle{box} = [draw=black!90, thick, rectangle, rounded corners,
                     inner sep=10pt, inner ysep=20pt, dotted
                  ]
\tikzstyle{title} = [draw=black!90, fill=black!5, semithick, top color=white,
                     bottom color = black!5, text=black!90, rectangle,
                     font=\small, inner sep=2pt, minimum height=1.3em,
                     top color=tyellow2!27, bottom color=tyellow2!27
                    ]
% block styles
\tikzstyle{feature} = [rectangle,font=\scriptsize,rounded corners=1mm,thick,%
                       draw=black!80, top color=tblue2!20,bottom color=tblue2!25,%
                       draw, minimum height=1.1em, text centered,%
                       inner sep=2pt%
                      ]
%\tikzstyle{pscore} = [rectangle,font=\scriptsize,rounded corners=1mm,thick,%
%                     draw=black!80, top color=tgreen3!20,bottom color=tgreen3!27,%
%                     draw, minimum height=1.1em, text centered,%
%                     inner sep=2pt%
%                    ]
%\tikzstyle{nscore} = [rectangle,font=\scriptsize,rounded corners=1mm,thick,%
%                     draw=black!80, top color=tred2!20,bottom color=tred2!25,%
%                     draw, minimum height=1.1em, text centered,%
%                     inner sep=2pt%
%                    ]
\tikzstyle{score} = [rectangle,font=\scriptsize,rounded corners=1mm,thick,%
                     draw=black!80, top color=tgreen3!10,bottom color=tgreen3!15,%
                     draw, minimum height=1.1em, text centered,%
                     inner sep=2pt%
                    ]
\begin{adjustbox}{center}
\setlength{\tabcolsep}{6pt}
\def\arraystretch{3}
\begin{tabular}{cc}

    \begin{tikzpicture}[node distance = 4.0em, auto]
        % creating a box with the title
        \node [box] (box) {%
        \begin{minipage}[t!]{0.776\textwidth}
            \vspace{0.9cm}\hspace{1.5cm}
        \end{minipage}
        };
        \node[title] at (box.north) {$\text{T}_\text{1}$};

        % feature node
        \node [feature] (feat1) at (0.08, 0.57) {sepal.length $<=$ 5.55?};
        \node [feature, below left  = 0.8em and -2.em of feat1] (feat2) {sepal.width $<=$ 2.75?};
        \node [feature, below right = 0.8em and -2.em of feat1] (feat3) {petal.width $<=$ 1.55?};

        % leaf nodes
        \node [score, below left  = 0.8em and -2.3em of feat2] (pos1) {Versicolor};
        \node [score, below right  = 0.8em and -2.5em of feat2] (neg1) {Setosa};
        \node [score, below left = 0.8em and -2.4em of feat3] (pos2) {Versicolor};
        \node [score, below right = 0.8em and -2.4em of feat3] (neg2) {Virginica};

        % arrows to the leaves
        \draw [->,thick,black!80] (feat1) to[] node[above, pos=1.2, font=\scriptsize] {yes} (feat2.north);
        \draw [->,thick,black!80] (feat1) to[] node[above, pos=1.1, font=\scriptsize] { no} (feat3.north);
        \draw [->,thick,black!80] (feat2) to[] node[above, pos=1.2, font=\scriptsize] {yes} (pos1.north);
        \draw [->,thick,black!80] (feat2) to[] node[above, pos=1.1, font=\scriptsize] { no} (neg1.north);
        \draw [->,thick,black!80] (feat3) to[] node[above, pos=1.2, font=\scriptsize] {yes} (pos2.north);
        \draw [->,thick,black!80] (feat3) to[] node[above, pos=1.1, font=\scriptsize] { no} (neg2.north);
    \end{tikzpicture}
    &
    \begin{tikzpicture}[node distance = 4.0em, auto]
        % creating a box with the title
        \node [box] (box) {%
        \begin{minipage}[t!]{0.545\textwidth}
            \vspace{0.9cm}\hspace{1.5cm}
        \end{minipage}
        };
        \node[title] at (box.north) {$\text{T}_\text{2}$};

        % feature node
        \node [feature] (feat1) at (-0.75, 0.57) {petal.width $<=$ 0.75?};
        \node [feature, below right = 0.8em and -2.2em of feat1] (feat2) {petal.length $<=$ 4.75?};

        % leaf nodes
        \node [score, below left  = 0.8em and -1.2em of feat1] (pos1) {Setosa};
        \node [score, below left = 0.8em and -2.4em of feat2] (pos2) {Versicolor};
        \node [score, below right = 0.8em and -2.4em of feat2] (neg) {Virginica};

        % arrows to the leaves
        \draw [->,thick,black!80] (feat1) to[] node[above, pos=1.2, font=\scriptsize] {yes} (pos1.north);
        \draw [->,thick,black!80] (feat1) to[] node[above, pos=1.1, font=\scriptsize] { no} (feat2.north);
        \draw [->,thick,black!80] (feat2) to[] node[above, pos=1.2, font=\scriptsize] {yes} (pos2.north);
        \draw [->,thick,black!80] (feat2) to[] node[above, pos=1.1, font=\scriptsize] { no} (neg.north);
    \end{tikzpicture}
    \\
    \multicolumn{2}{c}{
    \begin{tikzpicture}[node distance = 4.0em, auto]
        % creating a box with the title
        \node [box] (box) {%
        \begin{minipage}[t!]{0.54\textwidth}
            \vspace{0.9cm}\hspace{1.5cm}
        \end{minipage}
        };
        \node[title] at (box.north) {$\text{T}_\text{3}$};

        % feature node
        \node [feature] (feat1) at (-0.72, 0.57) {petal.width $<=$ 0.75?};
        \node [feature, below right = 0.8em and -2.2em of feat1] (feat2) {petal.width $<=$ 1.65?};

        % leaf nodes
        \node [score, below left  = 0.8em and -1.2em of feat1] (pos1) {Setosa};
        \node [score, below left = 0.8em and -2.4em of feat2] (pos2) {Versicolor};
        \node [score, below right = 0.8em and -2.4em of feat2] (neg) {Virginica};

        % arrows to the leaves
        \draw [->,thick,black!80] (feat1) to[] node[above, pos=1.2, font=\scriptsize] {yes} (pos1.north);
        \draw [->,thick,black!80] (feat1) to[] node[above, pos=1.1, font=\scriptsize] { no} (feat2.north);
        \draw [->,thick,black!80] (feat2) to[] node[above, pos=1.2, font=\scriptsize] {yes} (pos2.north);
        \draw [->,thick,black!80] (feat2) to[] node[above, pos=1.1, font=\scriptsize] { no} (neg.north);
    \end{tikzpicture} } %
  \end{tabular}
\end{adjustbox}

%% file: texfigs/ex2.tex
%------------------------------------------------------------------------------%
% File:        beacon.tex
%
% Description:
%
% Created:     7 Feb 2018.
%
% Author:      Alexey Ignatiev (aign).
%------------------------------------------------------------------------------%

% defining box structure
\tikzstyle{box} = [draw=black!90, thick, rectangle, rounded corners,
                     inner sep=10pt, inner ysep=20pt, dotted
                  ]
\tikzstyle{title} = [draw=black!90, fill=black!5, semithick, top color=white,
                     bottom color = black!5, text=black!90, rectangle,
                     font=\small, inner sep=2pt, minimum height=1.3em,
                     top color=tyellow2!27, bottom color=tyellow2!27
                    ]
% block styles
\tikzstyle{feature} = [rectangle,font=\scriptsize,rounded corners=1mm,thick,%
                       draw=black!80, top color=tblue2!20,bottom color=tblue2!25,%
                       draw, minimum height=1.1em, text centered,%
                       inner sep=2pt%
                      ]
\tikzstyle{pscore} = [rectangle,font=\scriptsize,rounded corners=1mm,thick,%
                     draw=black!80, top color=tgreen3!20,bottom color=tgreen3!27,%
                     draw, minimum height=1.1em, text centered,%
                     inner sep=2pt%
                    ]
\tikzstyle{scores} = [rectangle,font=\scriptsize,rounded corners=1mm,thick,%
                     draw=black!80, top color=tgreen3!10,bottom color=tgreen3!15,%
                     draw, minimum height=1.1em, text centered,%
                     inner sep=2pt%
                    ]
\tikzstyle{nscore} = [rectangle,font=\scriptsize,rounded corners=1mm,thick,%
                     draw=black!80, top color=tred2!20,bottom color=tred2!25,%
                     draw, minimum height=1.1em, text centered,%
                     inner sep=2pt%
                    ]

\begin{adjustbox}{center}
\setlength{\tabcolsep}{8pt}
\def\arraystretch{3}
\begin{tabular}{cc}
   \multicolumn{2}{c}{
    \begin{tikzpicture}[node distance = 4.0em, auto]
        % creating a box with the title
        \node [box] (box) {%
        \begin{minipage}[t!]{0.842\textwidth}
            \vspace{1.4cm}\hspace{1.5cm}
        \end{minipage}
        };
        \node[title] at (box.north) {$\text{T}_\text{1} $};

        % feature node
        \node [feature] (feat1) at (0.04, 0.9) {sepal.length $<=$ 5.55?};
        \node [feature, below left  = 0.8em and -0.2em of feat1] (feat2) {sepal.width $<=$ 2.75?};
        \node [feature, below right = 0.8em and -0.1em of feat1] (feat3) {petal.width $<=$ 1.55?};

        % leaf nodes
        \node [scores, below left  = 0.8em and -2.3em of feat2, align=left] (pos1)  { Setosa: 0.307 \\  {\bf Versicolor}: {\bf 0.615} \\ Virginica: 0.076};
        \node [scores, below right  = 0.8em and -2.5em of feat2, align=left] (neg1) { {\bf Setosa}: {\bf 1.0} \\  Versicolor: 0.0 \\  Virginica: 0.0};
        \node [scores, below left = 0.8em and -2.4em of feat3, align=left] (pos2) { Setosa: 0.093 \\  {\bf Versicolor}: {\bf 0.843} \\  Virginica: 0.062};
        \node [scores, below right = 0.8em and -2.4em of feat3, align=left] (neg2) { Setosa: 0.0 \\  Versicolor: 0.047 \\   {\bf Virginica}: {\bf 0.952}};

        % arrows to the leaves
        \draw [->,thick,black!80] (feat1) to[] node[above, pos=1.2, font=\scriptsize] {yes} (feat2.north);
        \draw [->,thick,black!80] (feat1) to[] node[above, pos=1.1, font=\scriptsize] { no} (feat3.north);
        \draw [->,thick,black!80] (feat2) to[] node[above, pos=1.2, font=\scriptsize] {yes} (pos1.north);
        \draw [->,thick,black!80] (feat2) to[] node[above, pos=1.1, font=\scriptsize] { no} (neg1.north);
        \draw [->,thick,black!80] (feat3) to[] node[above, pos=1.2, font=\scriptsize] {yes} (pos2.north);
        \draw [->,thick,black!80] (feat3) to[] node[above, pos=1.1, font=\scriptsize] { no} (neg2.north);
    \end{tikzpicture}
    } %
     \\
    \begin{tikzpicture}[node distance = 4.0em, auto]
        % creating a box with the title
        \node [box] (box) {%
        \begin{minipage}[t!]{0.553\textwidth}
            \vspace{1.4cm}\hspace{1.5cm}
        \end{minipage}
        };
        \node[title] at (box.north) {$\text{T}_\text{2}$ };

        % feature node
        \node [feature] (feat1) at (-0.73, 0.9) {petal.width $<=$ 0.75?};
        \node [feature, below right = 0.8em and -2.2em of feat1] (feat2) {petal.length $<=$ 4.75?};

        % leaf nodes
        \node [scores, below left  = 0.8em and -1.2em of feat1, align=left] (pos1) { {\bf Setosa}: {\bf 1.0} \\  Versicolor: 0.0  \\   Virginica: 0.0 };
        \node [scores, below left = 0.8em and -2.4em of feat2, align=left] (pos2) { Setosa: 0.0 \\   {\bf Versicolor}: {\bf 1.0}  \\  Virginica: 0.0 };
        \node [scores, below right = 0.8em and -2.4em of feat2, align=left] (neg) { Setosa: 0.0 \\   Versicolor: 0.166  \\  {\bf Virginica}: {\bf 0.833} };

        % arrows to the leaves
        \draw [->,thick,black!80] (feat1) to[] node[above, pos=1.2, font=\scriptsize] {yes} (pos1.north);
        \draw [->,thick,black!80] (feat1) to[] node[above, pos=1.1, font=\scriptsize] { no} (feat2.north);
        \draw [->,thick,black!80] (feat2) to[] node[above, pos=1.2, font=\scriptsize] {yes} (pos2.north);
        \draw [->,thick,black!80] (feat2) to[] node[above, pos=1.1, font=\scriptsize] { no} (neg.north);
    \end{tikzpicture}
    &
    \begin{tikzpicture}[node distance = 4.0em, auto]
        % creating a box with the title
        \node [box] (box) {%
        \begin{minipage}[t!]{0.56\textwidth}
            \vspace{1.4cm}\hspace{1.5cm}
        \end{minipage}
        };
        \node[title] at (box.north) {$\text{T}_\text{3}$};

        % feature node
        \node [feature] (feat1) at (-0.53, 0.9) {petal.width $<=$ 0.75?};
        \node [feature, below right = 0.8em and -2.2em of feat1] (feat2) {petal.width $<=$ 1.65?};

        % leaf nodes
        \node [scores, below left  = 0.8em and -0.5em of feat1, align=left] (pos1) { {\bf Setosa}: {\bf 1.0} \\ Versicolor: 0.0 \\ Virginica: 0.0};
        \node [scores, below left = 0.8em and -2.4em of feat2, align=left] (pos2) {Setosa: 0.0 \\  {\bf Versicolor}: {\bf 0.904} \\  Virginica: 0.095};
        \node [scores, below right = 0.8em and -2.4em of feat2, align=left] (neg) {Setosa: 0.0 \\  Versicolor: 0.05 \\ {\bf Virginica}: {\bf 0.95}};

        % arrows to the leaves
        \draw [->,thick,black!80] (feat1) to[] node[above, pos=1.2, font=\scriptsize] {yes} (pos1.north);
        \draw [->,thick,black!80] (feat1) to[] node[above, pos=1.1, font=\scriptsize] { no} (feat2.north);
        \draw [->,thick,black!80] (feat2) to[] node[above, pos=1.2, font=\scriptsize] {yes} (pos2.north);
        \draw [->,thick,black!80] (feat2) to[] node[above, pos=1.1, font=\scriptsize] { no} (neg.north);
    \end{tikzpicture}
\end{tabular}
\end{adjustbox}

%% file: texfigs/ex1.tex
%------------------------------------------------------------------------------%
% File:        beacon.tex
%
% Description:
%
% Created:     7 Feb 2018.
%
% Author:      Alexey Ignatiev (aign).
%------------------------------------------------------------------------------%

% defining box structure
\tikzstyle{box} = [draw=black!90, thick, rectangle, rounded corners,
                     inner sep=10pt, inner ysep=20pt, dotted
                  ]
\tikzstyle{title} = [draw=black!90, fill=black!5, semithick, top color=white,
                     bottom color = black!5, text=black!90, rectangle,
                     font=\small, inner sep=2pt, minimum height=1.3em,
                     top color=tyellow2!27, bottom color=tyellow2!27
                    ]
% block styles
\tikzstyle{feature} = [rectangle,font=\scriptsize,rounded corners=1mm,thick,%
                       draw=black!80, top color=tblue2!20,bottom color=tblue2!25,%
                       draw, minimum height=1.1em, text centered,%
                       inner sep=2pt%
                      ]
\tikzstyle{pscore} = [rectangle,font=\scriptsize,rounded corners=1mm,thick,%
                     draw=black!80, top color=tgreen3!20,bottom color=tgreen3!27,%
                     draw, minimum height=1.1em, text centered,%
                     inner sep=2pt%
                    ]
\tikzstyle{nscore} = [rectangle,font=\scriptsize,rounded corners=1mm,thick,%
                     draw=black!80, top color=tred2!20,bottom color=tred2!25,%
                     draw, minimum height=1.1em, text centered,%
                     inner sep=2pt%
                    ]
\tikzstyle{score} = [rectangle,font=\scriptsize,rounded corners=1mm,thick,%
                     draw=black!80, top color=tgreen3!10,bottom color=tgreen3!15,%
                     draw, minimum height=1.1em, text centered,%
                     inner sep=2pt%
                    ]  
\begin{adjustbox}{center}
\setlength{\tabcolsep}{16pt}
\def\arraystretch{3}
\begin{tabular}{ccc}
    \begin{tikzpicture}[node distance = 4.0em, auto]
        % creating a box with the title
        \node [box] (box) {%
        \begin{minipage}[t!]{0.154\textwidth}
            \vspace{0.9cm}\hspace{1.5cm}
        \end{minipage}
        };
        \node[title] at (box.north) {$\text{T}_\text{1}$ (setosa)};

        % feature node
        \node [feature] (feat) at (0, 0.25) {petal.length $<$ 2.45?};

        % leaf nodes
        \node [score, below left = 0.9em and -1.4em of feat] (pos)  { 0.42762};
        \node [score, below right = 0.9em and -1.6em of feat] (neg) {-0.21853};

        % arrows to the leaves
        \draw [->,thick,black!80] (feat) to[] node[above, pos=1.2, font=\scriptsize] {yes} (pos.north);
        \draw [->,thick,black!80] (feat) to[] node[above, pos=1.1, font=\scriptsize] { no} (neg.north);
    \end{tikzpicture}
    &
    \begin{tikzpicture}[node distance = 4.0em, auto]
        % creating a box with the title
        \node [box] (box) {%
        \begin{minipage}[t!]{0.264\textwidth}
            \vspace{0.9cm}\hspace{1.5cm}
        \end{minipage}
        };
        \node[title] at (box.north) {$\text{T}_\text{2}$ (versicolor)};

        % feature node
        \node [feature] (feat1) at (0.04, 0.57) {sepal.width $<$ 2.95?};
        \node [feature, below left  = 0.8em and -2.em of feat1] (feat2) {petal.width $<$ 1.7?};
        \node [feature, below right = 0.8em and -2.em of feat1] (feat3) {petal.length $<$ 3?};

        % leaf nodes
        \node [score, below left  = 0.8em and -2.3em of feat2] (pos1) {0.36131};
        \node [score, below right  = 0.8em and -2.5em of feat2] (neg1) {-0.18947};
        \node [score, below left = 0.8em and -2.4em of feat3] (pos2) {-0.21356};
        \node [score, below right = 0.8em and -2.4em of feat3] (neg2) {-0.03830};

        % arrows to the leaves
        \draw [->,thick,black!80] (feat1) to[] node[above, pos=1.2, font=\scriptsize] {yes} (feat2.north);
        \draw [->,thick,black!80] (feat1) to[] node[above, pos=1.1, font=\scriptsize] { no} (feat3.north);
        \draw [->,thick,black!80] (feat2) to[] node[above, pos=1.2, font=\scriptsize] {yes} (pos1.north);
        \draw [->,thick,black!80] (feat2) to[] node[above, pos=1.1, font=\scriptsize] { no} (neg1.north);
        \draw [->,thick,black!80] (feat3) to[] node[above, pos=1.2, font=\scriptsize] {yes} (pos2.north);
        \draw [->,thick,black!80] (feat3) to[] node[above, pos=1.1, font=\scriptsize] { no} (neg2.north);
    \end{tikzpicture}
    &
    \begin{tikzpicture}[node distance = 4.0em, auto]
        % creating a box with the title
        \node [box] (box) {%
        \begin{minipage}[t!]{0.215\textwidth}
            \vspace{0.9cm}\hspace{1.5cm}
        \end{minipage}
        };
        \node[title] at (box.north) {$\text{T}_\text{3}$ (virginica)};

        % feature node
        \node [feature] (feat1) at (-0.375, 0.57) {petal.length $<$ 4.75?};
        \node [feature, below right = 0.8em and -2.2em of feat1] (feat2) {petal.width $<$ 1.7?};

        % leaf nodes
        \node [score, below left  = 0.8em and -1.2em of feat1] (pos1) {-0.21869};
        \node [score, below left = 0.8em and -2.4em of feat2] (pos2) {0.08182};
        \node [score, below right = 0.8em and -2.4em of feat2] (neg) {0.42282};

        % arrows to the leaves
        \draw [->,thick,black!80] (feat1) to[] node[above, pos=1.2, font=\scriptsize] {yes} (pos1.north);
        \draw [->,thick,black!80] (feat1) to[] node[above, pos=1.1, font=\scriptsize] { no} (feat2.north);
        \draw [->,thick,black!80] (feat2) to[] node[above, pos=1.2, font=\scriptsize] {yes} (pos2.north);
        \draw [->,thick,black!80] (feat2) to[] node[above, pos=1.1, font=\scriptsize] { no} (neg.north);
    \end{tikzpicture}
    \\
    \begin{tikzpicture}[node distance = 4.0em, auto]
        % creating a box with the title
        \node [box] (box) {%
        \begin{minipage}[t!]{0.154\textwidth}
            \vspace{0.9cm}\hspace{1.5cm}
        \end{minipage}
        };
        \node[title] at (box.north) {$\text{T}_\text{4}$ (setosa)};

        % feature node
        \node [feature] (feat) at (0, 0.25) {petal.length $<$ 2.45?};

        % leaf nodes
        \node [score, below left = 0.9em and -1.4em of feat] (pos) {0.29522};
        \node [score, below right = 0.9em and -1.6em of feat] (neg) {-0.19674};

        % arrows to the leaves
        \draw [->,thick,black!80] (feat) to[] node[above, pos=1.2, font=\scriptsize] {yes} (pos.north);
        \draw [->,thick,black!80] (feat) to[] node[above, pos=1.1, font=\scriptsize] { no} (neg.north);
    \end{tikzpicture}
    &
    \begin{tikzpicture}[node distance = 4.0em, auto]
        % creating a box with the title
        \node [box] (box) {%
        \begin{minipage}[t!]{0.275\textwidth}
            \vspace{0.9cm}\hspace{1.5cm}
        \end{minipage}
        };
        \node[title] at (box.north) {$\text{T}_\text{5}$ (versicolor)};

        % feature node
        \node [feature] (feat1) at (0.125, 0.57) {sepal.width $<$ 2.95?};
        \node [feature, below left  = 0.8em and -2.em of feat1] (feat2) {petal.length $<$ 4.85?};
        \node [feature, below right = 0.8em and -2.em of feat1] (feat3) {petal.length $<$ 3?};

        % leaf nodes
        \node [score, below left  = 0.8em and -2.3em of feat2] (pos1) {0.27994};
        \node [score, below right  = 0.8em and -2.5em of feat2] (neg1) {-0.11330};
        \node [score, below left = 0.8em and -2.4em of feat3] (pos2) {-0.18999};
        \node [score, below right = 0.8em and -2.4em of feat3] (neg2) {-0.02829};

        % arrows to the leaves
        \draw [->,thick,black!80] (feat1) to[] node[above, pos=1.2, font=\scriptsize] {yes} (feat2.north);
        \draw [->,thick,black!80] (feat1) to[] node[above, pos=1.1, font=\scriptsize] { no} (feat3.north);
        \draw [->,thick,black!80] (feat2) to[] node[above, pos=1.2, font=\scriptsize] {yes} (pos1.north);
        \draw [->,thick,black!80] (feat2) to[] node[above, pos=1.1, font=\scriptsize] { no} (neg1.north);
        \draw [->,thick,black!80] (feat3) to[] node[above, pos=1.2, font=\scriptsize] {yes} (pos2.north);
        \draw [->,thick,black!80] (feat3) to[] node[above, pos=1.1, font=\scriptsize] { no} (neg2.north);
    \end{tikzpicture}
    &
    \begin{tikzpicture}[node distance = 4.0em, auto]
        % creating a box with the title
        \node [box] (box) {%
        \begin{minipage}[t!]{0.215\textwidth}
            \vspace{0.9cm}\hspace{1.5cm}
        \end{minipage}
        };
        \node[title] at (box.north) {$\text{T}_\text{6}$ (virginica)};

        % feature node
        \node [feature] (feat1) at (-0.375, 0.57) {petal.length $<$ 4.75?};
        \node [feature, below right = 0.8em and -2.2em of feat1] (feat2) {petal.width $<$ 1.7?};

        % leaf nodes
        \node [score, below left  = 0.8em and -1.2em of feat1] (pos1) {-0.19776};
        \node [score, below left = 0.8em and -2.4em of feat2] (pos2) {0.08067};
        \node [score, below right = 0.8em and -2.4em of feat2] (neg) {0.30170};

        % arrows to the leaves
        \draw [->,thick,black!80] (feat1) to[] node[above, pos=1.2, font=\scriptsize] {yes} (pos1.north);
        \draw [->,thick,black!80] (feat1) to[] node[above, pos=1.1, font=\scriptsize] { no} (feat2.north);
        \draw [->,thick,black!80] (feat2) to[] node[above, pos=1.2, font=\scriptsize] {yes} (pos2.north);
        \draw [->,thick,black!80] (feat2) to[] node[above, pos=1.1, font=\scriptsize] { no} (neg.north);
    \end{tikzpicture}
\end{tabular}
\end{adjustbox}

%------------------------------------------------------------------------------%

%% file: tenc.tex
\paragraph{Explanation encoding for TEs.}
We start by detailing how to encode the trees of a TE $\mfrk{T}$ 
before describing the classification function $\kappa$. 
This paper exploits the propositional encoding used in recent
work for computing formal explanations for  RFs and  BTs 
(XGBoost~\cite{guestrin-kdd16a})~\cite{ims-ijcai21,iisms-aaai22}.
To the best of our knowledge, there is no existing propositional 
encoding to explain RF with weighted voting. 
As a result, we aim to address this gap and present a generic 
MaxSAT encoding for TE capturing RFwv, RFmv and BT.
The encoding comprises: 1) the structure of a TE $\mfrk{T}$, and 2) the
weighted voting prediction $(\kappa(\mbf{x})\not=c)$.

Given a tree ensemble $\mfrk{T}$ defining classifier $\kappa$, 
for each feature $i \in \fml{F}$
we can define the set of \emph{split points} $S_i \subset \mbb{D}_i$ that
appear for that feature in all trees, i.e. $S_i = \{ d ~|~ x_i < d
\text{~appears in some tree~} T_1, \ldots T_n \}$.
Suppose $[s_{i,1}, s_{i,2}, \ldots, s_{i,|S_i|}]$ are the split points in
$S_i$ sorted in increasing order.
Using this we can define set of disjoint intervals for feature $i$:
$I^i_1 \equiv [\min(\mbb{D}_i),  s_{i,1})$,
$I^i_2 \equiv  [s_{i,1}, s_{i,2})$, \ldots,
$I^i_{|S_i|+1} \equiv [s_{i,|S_i|}, \max(\mbb{D}_i)]$
where $\mbb{D}_i = I^i_1 \cup I^i_2 \cup \cdots \cup I^i_{|S_i|+1}$.
The key property of a TE is that  
$\forall \mbf{x} \in I^1_{e_1} \times I^2_{e_2} \times \cdots \times I^m_{e_m}$ 
s.t.\  $e_i$ is defined as the interval index for feature $i$ which includes
the value of $\mbf{v}$, i.e. $v_i \in I^i_{e_i}$, then we have   
$\kappa(\mbf{x}) =  \kappa(\mbf{v}) = c$.
That is we only need to reason about intervals rather than particular values
to explain the behaviour of a tree ensemble.

We define Boolean variables $\bb{x_i < d}, d \in S_i$ for each feature 
$i\in \fml{F}$. 
With this we can define a Boolean variable $\bb{R_l}$ for each path $R_l \in
\mfrk{T}$ defined as
$$
\bb{R_l} \leftrightarrow \left( \bigwedge\nolimits_{(x_i < d) \in R_l} \bb{x_i < d} \wedge
\bigwedge\nolimits_{\neg (x_i < d) \in R_l} \neg \bb{x_i < d} \right)$$

The total weights of a class $c$ can be determined by 
$\sum_c=\sum_{T_{t}\in\mfrk{T}}{\sum_{\bb{R_l}\in T_{t}, cl(l) = c}{w_l\cdot \bb{R_l} }}$.
Recall that the goal is to reason about $\mfrk{T}$ to determine whether 
there exist a counter-example $\mbf{x}$, 
$\kappa(\mbf{x}) \neq c$, that is consistent with $\fml{X}\subset\fml{F}$ (i.e.\ 
$\bigland\nolimits_{i\in\fml{X}} (x_i \in I^i_{e_i})$).
This can be achieved by verifying  $\sum_{c'} > \sum_c$ where $c$ 
is the target class and $c' \in \fml{K}, c' \neq c$ are the opponent 
classes.
We formulate this in the {\it weighted partial} MaxSAT formula as 
a set of soft clauses of the form 
\begin{align}
\bigland\nolimits_{T_{t}\in\mfrk{T}}  & \left[
\bigland\nolimits_{\bb{R_l}\in T_{t}, cl(l) = c'} (\bb{R_l}, w_l) \right. \wedge 
\nonumber \\
&\left. \:\:  \bigland\nolimits_{\bb{R_l}\in T_{t}, cl(l) = c} (\bb{R_l}, -w_l)  \right] \nonumber
\end{align}
which defines the objective function to maximize:   
$\fml{S}_{c,c'}=\sum\nolimits_{c'} - \sum\nolimits_c$ for each
formula checking $(\kappa(\mbf{x}) = c' ) \neq c$ 
for a class $c'$ against the target $c$.
Note that it is not required to search for the optimal solution 
for $\fml{S}_{c,c'}$ since we are only interested in identifying 
a class $c'$ that would have a greater score that $c$.%\footnote{%
%In our experiments, we tune the MaxSAT oracle to return the first 
%valid solution met instead of the optimal one.}

We emphasize that the described TE encoding is dissociated from  
the most general explanations encodings, presented next.

%% file: mxp.tex
\section{Most General Explanations}
Let us assume a tree ensemble $\mfrk{T}$ and features $i\in\fml{F}$ are
numerical. Hence, for each feature we obtain a set of intervals $\set{I_{1}, \dots, I_{n}}$
such that $\left(\bigcup_{j=1}^{n} I_j \right)  = \mbb{D}_i$.
Assuming the domain $\mbb{D}_i$ is finite, we can define the \emph{proportion size} of an interval $I_j$ for feature
$i$ as $prop_i(I_j) = \nicefrac{|I_j|}{|\mbb{D}_i|}$.  For infinite domains
$\mbb{D}_i$ we treat its size as $|\mbb{D}_i| = \max \{v_i ~|~ \mathbf{v} \in \mbb{T}\}
- \min \{ v_i ~|~ \mathbf{v} \in \mbb{T}\}$
where $\mbb{T}$ is the training
set used to train the TE.
We can alternatively define the \emph{data proportion size} of an interval
$I_j$ for feature $i$ as $data_i(I_j) = \nicefrac{| \{ \textbf{v} \in \mbb{T} ~|~ v_i
\in I_j \} |}{| \mbb{T} |}$ which give the proportion of the training data
that appears in the interval.
We can extend both size definitions to work on interval $\mbb{E}_i$ for feature
$i$ in the natural manner: $prop_i(\mbb{E}_i) = \sum_{I_j \in \mbb{E}_i}
prop_i(I_j)$,
and $data_i(\mbb{E}_i) = \sum_{I_j \in \mbb{E}_i}
data_i(I_j)$.

We can now compute the size of a region $\mbb{E}$ covered by an
inflated explanation $(\fml{X}, \mbb{E})$ as either $prop(\mbb{E}) =
\Pi_{i \in  \fml{F}} prop_i(\mbb{E}_i)$ or $data(\mbb{E}) = \Pi_{i \in
\fml{F}} data_i(\mbb{E}_i)$.
Either of these can be used to define the size measure $s(\mbb{E})$.

Note that in the context of explanation inflation, excluding a feature
$i\in\fml{F}$ from an explanation $(\fml{X},\mbb{E})$ can be seen as
either removing it from $\fml{X}$ or, equivalently, as setting
$\mbb{E}_i=\mbb{D}_i$.
For this reason, hereinafter we denote \iaxp's as $(\fml{X},\mbb{E})$
or $\mbb{E}$ interchangeably.
In order to compute maximal size explanations, it will be convenient
to compute instead the logarithm of the size of an explanation, thus
replacing product by sum.
Namely, We define the \emph{feature space coverage} of an explanation
$\mbb{E}$, $FSC_s(\mbb{E})$ as the log of the size of the explanation
using size function $s$:
\begin{align} \label{eq:fsc}
    FSC_s(\mbb{E})  = \: \log  \prod\nolimits_{i\in\fml{F}} s(\mbb{E}_i)
    			= \:  \sum\nolimits_{i\in\fml{F}} \left( \log s(\mbb{E}_i) \right)  %
\end{align}

One key observation is that (\ref{eq:fsc}) assumes a uniform distribution 
over $\mbb{E}$. However,  we emphasize that the (training) data 
distribution can be seamlessly incorporated into $FSC_s$ by weighting 
the interval scores $\mbb{E}_i$ according to their probability distribution.

\paragraph{Maximum inflated explanation (\mxiaxp).}
Given an explanation problem $(\fml{M}, (\mathbf{v}, c))$ and a size
metric $s$, a \emph{maximum inflated abductive explanation}
$(\fml{X},\mbb{E})$ defines the intervals for all the features
$i\in\fml{X}$, such that the following conditions hold,
%%where set of subsets of domains, one for each $i\in\fml{X}$, such that
%
\begin{align}
  \label{eq:mxiaxp1}
  %
%  \forall(i \in \fml{F}).\,\, v_i \in \mbb{E}_i  \\
  (\fml{X},\mbb{E}) \in \mbb{A}(\fml{E}) \\
  % \forall(\mbf{x}\in\mbb{F}).%
  % \left[\bigland\nolimits_{i\in\fml{F}}(x_i\in\mbb{E}_i)\right]%
  % \limply(\kappa(\mbf{x})=c)  \\
  \label{eq:mxiaxp2}
  \forall(\mbb{E}^\prime\subseteq\mbb{F}).
    (\fml{X},\mbb{E}^\prime)\not\in\mbb{A}(\fml{E}) \lor FSC_s(\mbb{E}^\prime) \le FCS_s(\mbb{E})
  % \forall (\mbb{E}^\prime \subset \mbb{F}).
  % \left[
  %   \bigland\nolimits_{i\in\fml{F}} v_i \in \mbb{E}^{\prime}_i \right]
  % \wedge \nonumber \\
  %   \left( \forall(\mbf{x}\in\mbb{F}). \left[
  %     \bigland\nolimits_{i\in\fml{F}}  x_i \in  \mbb{E}^{\prime}_i \right]
  %   \rightarrow
  %   (\kappa(\mbf{x}) = c) \right) \nonumber \\ %
  % 	  \limply FSC_s(\mbb{E}^\prime) \le FCS_s(\mbb{E})
\end{align}
In other words, $(\fml{X}, \mbb{E})$ is a correct inflated AXp for the
explanation problem $\fml{E}$, and there is no other correct inflated
AXp, which has a region $\mbb{E}^{\prime}$ with a larger size under
$FSC_s$.

\paragraph{Implicit hitting dualization algorithm.}
We devise an algorithm inspired by the \emph{implicit hitting set
dualization}
paradigm~\cite{stuckey-padl05,karp-soda11,bacchus-cp11,lm-cpaior13,iplms-cp15,jarvisalo-kr16}
%~\cite{stuckey-padl05,karp-soda11,bacchus-cp11,pms-aaai13,lm-cpaior13,iplms-cp15,jarvisalo-kr16}
for computing maximum inflated AXps.
The idea is to use an oracle that searches for explanation candidates
$\mbb{E}$ and a second oracle that checks if $\mbb{E}$ is an \iaxp.
The latter solver deals with the encoding of the prediction function
of the TE and checks if \eqref{eq:iaxp} holds for the
candidate $\mbb{E}$ (in which case it is indeed an \iaxp), and reports
a counterexample otherwise.
The former oracle, i.e., guessing candidate $\mbb{E}$, encodes a
MaxSAT problem where the objective function to maximize is
\eqref{eq:fsc} and the hard constraints are representing the intervals
$\mbb{E}_i$.
Next, we will describe in detail the encoding that allows computing a
minimum hitting set $\mbb{E}$ of all \icxp's maximizing
$FSC_s(\mbb{E})$.

At each iteration of the algorithm, we obtain a range $\mbb{E}$ where
for each feature $i$, $\mbb{E}_i \subseteq \mbb{D}_i$, $\mbb{E}_i =
\bigcup_{l \leq j \leq u} I^i_j$ 
 ( s.t. $I^i_j \equiv  [s_{i,l}, s_{i,u})$
 %$l_i$ and $u_i$ are lower and upper bound indices of interval $I^i_j$ 
 )   
and $v_i \in \mbb{E}_i$.
We construct $\mbb{E}$ and check if it is an \iaxp, i.e. there is
no adversarial example.

Observe that it is sufficient to show $\mbb{E}$ is not a Weak \iaxp
to prove the existence of a CXp in $\mbb{E}$  --- \eqref{eq:icxp}
holds --- by using any encoding of the TE to find
a solution in $\mbf{x} \in \mbb{E}$ where $\kappa(\mbf{x}) \neq c$.
This is achieved by fixing the Booleans of the TE encoding implied
by the candidate explanation $\mbb{E}$, i.e.
$\bb{x_i < d} = \top$ when $d \leq \msf{sup}(\mbb{E}_i)$ and
$\bb{x_i < d} = \bot$ when $d \geq \msf{inf}(\mbb{E}_i)$,
and maximizing $\fml{S}_{c,c'}$ for all $c' \neq c$.
If we find a solution with $\fml{S}_{c,c'} > 0$ this is an adversarial
example.
If the oracle encoding the TE decides \eqref{eq:iaxp} to be
false, that is it finds an adversarial example, then we reduce the
adversarial example into a CXp $\fml{Y}\subseteq\fml{F}$ (or inflated
CXp $(\fml{Y}, \mbb{G}$) and block it in the hitting set
oracle such that the next candidate hits this CXp.
Otherwise, the oracle reports \eqref{eq:iaxp} to be true, subsequently
we conclude that $\mbb{E}$ meets the conditions
\eqref{eq:mxiaxp1}--\eqref{eq:mxiaxp2}.
Pseudo-code for the algorithm is shown in \cref{alg:mxiaxp}.

\begin{algorithm}[t]
\input{./algs/mxiaxp}
  \caption{Computing maximum inflated AXp for TE} \label{alg:mxiaxp}
\end{algorithm}

A consequence of the MHS duality between
\iaxp's and \icxp's~\cite{iisms-aaai24} is that the checker oracle
reports a new candidate to be an \iaxp and the algorithm terminates
before the hitting set oracle exhausts hitting sets to enumerate.
Moreover, it is clear that the hitting set oracle finds the largest
unblocked interval for each iteration that maximizes the score of the
objective function $FSC_s(\mbb{E})$.
Also, note that at the first iteration, as the oracle does not know
anything about the model (no blocked subdomains yet), it will cover
all the feature space (all features are free).
Finally, as a simple improvement of the approach, one can compute
initially  iCXp's of size 1 and block those adversarial regions
at the beginning, before entering the hitting set enumeration loop.

\input{./tabs/varoles}
Generally, as \autoref{alg:mxiaxp} exploits the ideas of implicit
hitting set dualization, it shares the worst-case exponential number
of iterations with the rest of this paradigm's instantiations.
Note that the number of iterations here grows exponentially not only
with respect to the number of features as in the standard
smallest-size AXp extraction~\cite{inms-aaai19,inams-aiia20} but also
on the feature domain size.

\paragraph{Naive MaxSAT encoding.}
What remains to explain is how we encode the candidate enumeration process as
a MaxSAT problem $\langle \fml{H}, \fml{S} \rangle$,
and how we create the blocking clauses.

We create Boolean variables $y^i_{l,u}$ to represent the interval $\mbb{E}_i
= \cup_{l \leq j \leq u} I^i_j$. Note that we can restrict to cases where $l \leq
e_i \leq u$ since any \iaxp must include the interval of the example being
explained $e_i$ for each feature $i$.
Importantly, we can define the log of the weight of each interval
$w^i_{l,u} = \log (\sum_{l \leq j \leq u} s(I^i_k))$.
In order to use MaxSAT we need to fix the entire interval for feature $i$
so that the resulting objective is a sum. Hence we generate $O(|S_i|^2)$
Booleans for each feature $i$.

The only hard clauses $\fml{H}$ of the MaxSAT model encode the cardinality constraints
\begin{equation}
\forall_{i \in \fml{F}} \sum\nolimits_{1\le l\le e_i \leq u \le|S_i|+1}{y^i_{l,u}}
\le 1 \label{eq:card}
\end{equation}
enforcing that for each feature at most one interval covering the
instance to be explained is selected.
The soft clauses $\fml{S}$ of the MaxSat model
are $(y^i_{l,u}, w^i_{l,u})$, indicating our
objective is to maximise $\sum_{i \in \fml{F}} w^i_{l,u} y^i_{l,u} =
FSC_s(\mbb{E})$.

\begin{example}\label{ex:naive}
Consider the feature $i = 4$, $petal.width$ of the RF of
%Figure~\ref{fig:RF}(a), 
\cref{fig:RFmv}, then $S_4 = \{ 0.75, 1.55, 1.65 \}$
and we assume $\mbb{D}_4 = [0,3]$. This defines 4 intervals
$I^4_1 = [0,0.75)$, $I^4_2 = [0.75,1.55)$, $I^4_3 = [1.55,1.65)$,
$I^4_4 = [1.65,3]$. If we are explaining the instance from
Example~\ref{ex:RF(mv)} then the interval where the instance sits is
$e_4 = 1$ since $petal.width = 0.2$.
We construct Booleans $y^4_{1,1}$, $y^4_{1,2}$, $y^4_{1,3}$, $y^4_{1,4}$
representing $\mbb{E}_4$ as one of $[0,0.75)$, $[0,1.55)$, $[0,1.65)$, $[0,3]$.
We require at most one $y^4_{l,u}$ to be true, and give them weights
defined by the interval weight $FSC_s(I^4_{l,u})$.
\end{example}

A solution to the MaxSAT model assigns exactly one interval to each
feature, thus defining a candidate explanation $\mbb{E}$.
Optimality of the MaxSAT solution means it will be the candidate
explanation of largest total size.

We use the TE MaxSAT model  to determine if $\mbb{E}$ include a counterexample.
If this fails then $\mbb{E}$ is an \iaxp, and by the maximality of size
it is the maximum inflated AXp.
Otherwise, we have a counterexample which we need to exclude from future
intervals.

Assume the TE model, given a candidate explanation
$\mbb{E}$ returns a counterexample, then we have a solution on the $\bb{x_i
  < d}$ variables, falling within $\mbb{E}$ where $\kappa(\mbf{w}) \neq c$
for all $\mbf{w}$ which satisfy these bounds.
We can extract a counterexample range $\mbb{G}$ from the solution,
where $\mbb{G}_i = I^i_{a_i,b_i}$ given by $a_i =
\max(\{ j+1 ~|~ j \in 1..|S_i|, \bb{x_i < s_{i,j}} = \bot
  \} \cup \{ 1 \})$ and
$b_i = \min(\{ j ~|~ j \in 1..|S_i| ~|~ \bb{x_i < s_{i,j}} = \top \} \cup \{ |S_i|+1 \})$.
We now add clauses to the MaxSAT model to prevent selecting candidate
explanations that would cover this adversarial example.
Note that if $a_i \leq e_i \leq  b_i$ then we cannot differentiate the
counterexample from the instance being explained using feature $i$.
Hence we want to express the constraint that $\mbb{E}_i \not\supseteq
I^i_{a_i,b_i}$ for at least one $i \in \fml{F}$ where it is not the case that
$a_i \leq e_i \leq b_i$.

We can express the condition $\mbb{E}_i \not\supseteq
I^i_{a_i,b_i}$ as the conjunction
$\bigwedge_{l\leq a_i \wedge b_i \leq u} \neg y^i_{l,u}$, since we can no longer choose an
interval, which covers the interval of the counterexample.
The ``blocking clause'' we add to the MaxSAT model is then the disjunction
$\bigvee_{i \in \fml{Y}, \neg (a_i \leq e_i \leq b_i)} \bigwedge_{l\leq a_i \wedge
  b_i\leq u} \neg y^i_{l,u}$.
To translate this into clausal form, we add auxiliary Boolean variables $\bb{\mbb{E}_i
  \not\supseteq I^i_{a_i,b_i}}$ and the clausal encoding of:
$ \bb{\mbb{E}_i \not\supseteq I^i_{a_i,b_i}} \leftrightarrow \bigwedge_{l\leq a_i \wedge b_i \leq u} \neg y^i_{l,u}$
and
$\bigvee_{i \in \fml{Y}, \neg (a_i \leq e_i \leq b_i)} \bb{\mbb{E}_i   \not\supseteq I^i_{a_i,b_i}}$
  to the MaxSAT model.

\begin{example}\label{ex:block}
Suppose for explaining the instance of Example~\ref{ex:RF(mv)} we
find counterexample $\mbb{G}$ corresponding to sample $\mbf{w} = (6.0, 3.5,
1.4, 0.8)$ which is predicted as Versicolor and
only differs in the $petal.width$ value. For this example
$\bb{x_4 < 0.75} = \bot$ and $\bb{x_4 < 1.55} = \top$, so
$\mbb{G}_4 = I^4_{2,2}$. We enforce new variable
$\bb{\mbb{E}_4 \not\supseteq I^4_{2,2}} \leftrightarrow (\neg y^4_{1,2} \wedge \neg y^4_{1,3} \wedge \neg
y^4_{1,4})$,
and add the blocking clause $\bb{\mbb{E}_4 \not\supseteq I^4_{2,2}}$ forcing
it to hold. We can no longer choose these intervals when searching for \iaxp
candidate $\mbb{E}$.
\end{example}

Note that we only add definitions for Booleans $\bb{\mbb{E}_i
  \not\supseteq I^i_{a_i,b_i}}$ \emph{on demand}, that is when they occur in a
counterexample.  This prevents us creating a large initial model, where many
of these auxiliary Booleans may never be required.

\paragraph{Bounds-based MaxSAT encoding.}

The above MaxSAT encoding creates large cardinality encodings
\eqref{eq:card},
and may generate many Booleans of the form $\bb{\mbb{E}_i \not\supseteq
  I^i_{a_i,b_i}}$ in order to block counterexamples.
We can do this much more succinctly using a bounds representation.

Here we introduce Boolean variables to represent the lower and upper bounds
of the interval $\mbb{E}_i$: $\{ \bb{l_i \geq d} ~|~ d \in S_i, d \leq \min(I^i_{e_i}) \}$
and $\{ \bb{u_i < d} ~|~ d \in S_i, d \geq \max(I^i_{e_i}) \}$. Note that we do not need to represent
bounds which exclude the explained instance value for this feature $i$,
$v_i$.
We add to $\fml{H}$ the implications
$\bb{l_i \geq s_{i,j+1}} \rightarrow \bb{l_i \geq s_{i,j}}, 1 \leq j < e_i$
and
$\bb{u_i < s_{i,j}} \rightarrow \bb{u_i < s_{i,j+1}}, e_i \leq j \leq |S_i|$.

We define the interval variables $y^i_{l,u}$ using clauses in $\fml{H}$ encoding
$$
\begin{array}{rcl}
   y^i_{l,u} & \leftrightarrow & \bb{l_i \geq s_{i,l-1}} \wedge \neg \bb{l_i \geq
    s_{i,l}} \wedge \\
	     & & \neg \bb{u_i < s_{i,u}} \wedge \bb{u_i < s_{i,u+1}}
\end{array}
$$
where $\bb{l_i \geq s_{i,0}}$ and $\bb{u_i < s_{i,|S_i|+1}}$ are both treated
as $\top$.
We use the same soft clauses to define the objective as in the Naive encoding.

\begin{example}
Encoding feature $petal.width$ we generate only the bounds
$\bb{u_4 < 0.75}$, $\bb{u_4 < 1.55}$, $\bb{u_4 < 1.65}$ and
implications $\bb{u_4 < 0.75} \rightarrow \bb{u_4 < 1.55}$ and
$\bb{u_4 < 1.55} \rightarrow \bb{u_4 < 1.65}$, and
$y^4_{1,1} \leftrightarrow \bb{u_4 < 0.75}$,
$y^4_{1,2} \leftrightarrow \neg \bb{u_4 < 0.75} \wedge \bb{u_4 < 1.55}$,
$y^4_{1,3} \leftrightarrow \neg \bb{u_4 < 1.55} \wedge \bb{u_4 < 1.65}$,
and $y^4_{1,4} \leftrightarrow \neg \bb{u_4 < 1.65}$.
\end{example}

If the test whether $\mbb{E}$ is an \iaxp fails, we are given an (inflated)
CXp $(\fml{Y}, \mbb{G})$. Adding a blocking clause to the bounds-based
MaxSAT encoding is much simpler.
Suppose $\mbb{G}_i = I^i_{a_i,b_i}$ where $e_i \not\in [a_i,b_i]$ then
if $e_i < a_i$ we can hit the \icxp for feature $i$ by enforcing $\bb{u_i
  \leq s_{i,a_i-1}}$, similarly
if $e_i > b_i$ we can hit the \icxp for feature $i$ by enforcing
$\bb{l_i \geq s_{i,b_i}}$.
The blocking clause is the disjunction of these literals for
all $i \in \fml{F}$ where $e_i \not\in [a_i,b_i]$.

\begin{example}
Given the same counter example as in Example~\ref{ex:block}
we have $a_4 = b_4 = 2$ and $e_4 = 1$, so
the blocking clause we add for the bounds model is simply
$\bb{u_4 < 0.75}$.
\end{example}

\paragraph{MIP encoding.}
We can easily rewrite the bounds based MaxSAT encoding to a mixed
integer linear program (MIP model).
All the Boolean variables become 0-1-integer variables.
The implications become simple inequalities, e.g. $\bb{l_i \geq
s_{i,j+1}} \rightarrow \bb{l_i \geq s_{i,j}}$ becomes $\bb{l_i \geq
s_{i,j+1}} \leq \bb{l_i \geq s_{i,j}}$.
The definitions of the interval variables $y^i_{l,u}$ become a series
of inequalities where negation $\neg l$ is modelled by $1-l$.
For example the definition of $y^i_{l,u}$ becomes
$$
\begin{array}{rcl}
y^i_{l,u} & \leq & \bb{l_i \geq s_{i,l-1}} \\
y^i_{l,u} & \leq & 1 - \bb{l_i \geq s_{i,l}} \\
y^i_{l,u} & \leq & 1 - \bb{u_i < s_{i,u}} \\
y^i_{l,u} & \leq & \bb{l_i \geq s_{i,u+1}} \\
y^i_{l,u} + 3 & \geq &
\bb{l_i \geq s_{i,l-1}} + 1 - \bb{l_i \geq s_{i,l}} +  \\
&& 1 - \bb{u_i < s_{i,u}}
+ \bb{l_i \geq s_{i,u+1}}
\end{array}
$$
The objective in the MIP model is to maximize $\sum_{i \in \fml{F},1 \leq
l\leq e_i \leq u \leq |S_i|+1} w^i_{l,u} \cdot y^i_{l,u}$, which is
completely analogous to the MaxSAT encoding.
The blocking clauses are also representable as a simple linear inequality.

%% file: algs/mxiaxp.tex
\begin{flushleft}
\hspace*{\algorithmicindent}
\textbf{Input}:  {Expl. prob. $\fml{E} = (\fml{M},(\mbf{v},c))$; 
		WCNF $\langle\fml{H},\fml{S}\rangle$} \\
\hspace*{\algorithmicindent}
\textbf{Output}: {One \mxiaxp $(\fml{X},  \mbb{E})$}
\end{flushleft}
	
\begin{algorithmic}[1]
  \Repeat
  \State{$(\mu,\mbf{s})\gets\MxSAT(\fml{H},\fml{S})$}
  \State{$\mbb{E}\gets\{ \bigcup_{l \leq j \leq u} I^i_j  \,|\,  \forall(y^i_{l,u}\in\fml{S}). \mu(y^i_{l,u})=1\}$}
  \State{$\fml{X}\gets\{i\in\fml{F}\,|\, |\mbb{E}_i| < |\mbb{D}_i| \}$}  
  \State{$\outc\gets\neg\msf{WiAXp}(\fml{E}; \fml{X},\mbb{E})$ } 
  \If{$\outc=\True$}
  \State{$(\fml{Y}, \mbb{G})\gets \msf{FindiCXp}(\fml{E}; \fml{F}, \mbb{E})$}
  \State{$\fml{H}\gets\fml{H}\cup \msf{newBlockCl}(\fml{Y}, \mbb{G})$}
  \EndIf
  \Until{$\outc=\False$}
  \State{\Return $(\fml{X},  \mbb{E})$}
\end{algorithmic}

%% file: tabs/varoles.tex
\begin{table}[ht]
\centering
\resizebox{1.0\columnwidth}{!}{
\begin{tabular}{cl}
\toprule
Notation &  Role \\
\midrule
$y^i_{l,u}$ & Boolean variable encoding an interval $\mbb{E}_i= \cup_{l \leq j \leq u} I^i_j$ of feature $i$ \\
$\bb{\mbb{E}_i \not\supseteq I^i_{a_i,b_i}}$ & Boolean variable to test  if a sub-interval $I^i_{a_i,b_i}$  is (not) in $\mbb{E}_i$ \\
$\bb{l_i \geq s_{i,j}}$ & Boolean variable encoding the lower bound of an interval $\mbb{E}_i$ \\ %(s.t. $s_{i,j}$ is a split point generated by $\mfrk{T}$)
$\bb{u_i \le s_{i,j}}$ & Boolean variable encoding the upper bound of an interval $\mbb{E}_i$  \\
\midrule
$w^i_{l,u}$ & Denotes the weight of an interval $\mbb{E}_i=\cup_{l \leq j \leq u} I^i_j$ of feature $i$ \\
$FSC_s(\mbb{E})$ & Explanation coverage function $FSC_s(\mbb{E}) = \sum_{i \in \fml{F}} w^i_{l,u} y^i_{l,u}$ \\
 			      & to maximize \\
\bottomrule
\end{tabular} }
\caption{\footnotesize{%
Outline of variables/notations declared in the encodings.}}
\label{tab:notations}
\end{table}

%% file: res.tex
\section{Experiments} \label{sec:res}
This section presents a summary of empirical assessment
of computing  maximum inflated abductive explanations
for tree ensembles --- the case study of RFmv and BT ---  trained on some
of the widely studied datasets.

The empirical evaluation also includes the assessment of 
RFwv encoded with our MaxSAT encoding. 
%

%Our evaluation aims to investigate the following 
%research questions:
%
%\begin{itemize}
%%
% \item  \textbf{RQ1}: Are \emph{hypercubes} representing \mxiaxp explanations much larger than \iaxp's on real world benchmarks?
% \item \textbf{RQ2}:  Are the proposed logical encodings scale for practical RFs and BTs?
% \item \textbf{RQ3}: Does our algorithm converge quickly to deliver the optimal explanation? 
%\end{itemize}

%
\input{tabs/RFs-iaxp}

\subsection{Experimental Setting}
%%\paragraph{Experimental setup.}
%
The experiments are conducted on Intel Core~i5-10500 3.1GHz CPU
with 16GByte RAM running Ubuntu 22.04 LTS.
A time limit for each tested instance is fixed to 900 seconds
(i.e.\ 15 minutes); whilst the memory limit is set to 4 GByte.

%%\paragraph{Benchmarks.}
%
The assessment of TEs (RFs and BTs) is performed on a selection of  publicly 
available datasets, which originate from UCI Machine Learning 
Repository~\cite{uci} and Penn Machine Learning 
Benchmarks~\cite{Olson2017PMLB} --- in total 12 datasets.
Note that datasets are
fully numerical as we are interested in assessing the proposed FSC
coverage metric, which discriminate the intervals given their size.
For categorical data, the FSC scores  are the same for all
domain values, then it is pointless in our empirical evaluation to
include them.
When training RFs, we used Scikit-learn~\cite{scikitlearn} and 
for BTs we applied XGBoost algorithm~\cite{yu-tc20}.

%%\paragraph{Prototype implementation.}
%
The proposed approach is %prototyped as a set of Python scripts 
implemented in \texttt{RFxpl}\footnote{\url{https://github.com/izzayacine/RFxpl}} 
and \texttt{XReason}\footnote{\url{https://github.com/alexeyignatiev/xreason}} 
Python packages. %\footnote{%
%The code of RF implementation is available at \url{https://github.com/izzayacine/RFxpl} 
%and BT at \url{https://github.com/alexeyignatiev/xreason}}. 
%
(We emphasize that our propositional encoding for RFwv has been publicly 
available in \texttt{RFxpl} on GitHub since November 2022 --- verifiable via 
the repository history --- and serves as a timestamp to guard against 
potential scooping.)
The \texttt{PySAT} toolkit\footnote{\url{https://pysathq.github.io}}~\cite{imms-sat18,itk-sat24}  
is used to instrument SAT or/and
MaxSAT oracle calls.
RC2~\cite{imms-jsat19} MaxSAT solver, which is  implemented
in PySAT, is applied to all MaxSAT encodings (for the TE operation and
hitting set dualization).
Moreover, Gurobi~\cite{gurobi} is applied to instrument MIP oracle calls
using its Python interface.
%
% The outlined propositional encoding for RFwv is implemented in Python
% as well, and built on PySAT library.

\subsection{\mxiaxp Results}
%
%\paragraph{RQ1.}
%
\cref{tab:RFs-iaxp} shows the coverage size of \mxiaxp and
baseline \iaxp explanations, for the case study of RFmv tree ensembles.
For a more fine-grained evaluation, we compare the scalability of the proposed
logical encodings, i.e.\ naive and bound-based MaxSAT encodings as well as
MIP encoding in \cref{tab:RFs-iaxp}.
As can be observed from~\cref{tab:RFs-iaxp}, the average domain coverage
of \mxiaxp's is at least twice and up to 10 times, larger than the average
coverage of \iaxp's for a half of datasets while it is fairly superior
for the remaining datasets.

Interestingly, we observe that the maximum ratio between feature space
coverage offered by \mxiaxp and
\iaxp  can be extremely high for the majority
of benchmarks (e.g.\ 3128 to 6048 times wider in {\it appendicitis},
{\it ecoli}, {\it bupa} and {\it new-thyroid})
with a few exceptions on smaller datasets (e.g. {\it breast-cancer}, {\it car} and
{\it iris}, resp.\ showing a ratio of 5.5, 4.1 and 2.3).
Furthermore, the average lengths of \iaxp and \mxiaxp remain 
consistently close, even when there is a large gap in coverage scores, 
and very succinct w.r.t. input data (e.g. $\sim 6.6\%$ in \emph{ann-thyroid}).  

\input{tabs/BTs-iaxp}
%  

%\paragraph{RQ2 \& RQ3.}
%
Performance-wise, we observe a substantial advantage of MIP oracle
(Gurobi) over MaxSAT (RC2)
despite the fact they use essentially the same
formulation of the problem.
Although there may be various reasons for this phenomenon,
the hitting set problem seems to be more amenable to MIP
solvers as they can take advantage of multiple efficient optimization
procedures, including linear relaxation, branch-and-bound methods
augmented with cutting planes resolution as well as both primal and
dual reasoning.
On the contrary, RC2 is a core-guided MaxSAT solver designed to handle
optimization problems that are inherently Boolean, which is the case
for the TE encoding we used in our work where RC2 shines.
As the results demonstrate, our approach empowered with Gurobi is able
to deliver explanations within $\sim$33 sec for the average runtime
for all the considered datasets (and a min. (resp.\ max.) of 0.14 sec
($\sim$3 min)), and able to terminate on all tests, whilst RC2 gets
timed out on 4 datasets (particularly for {\it wine-recog} in 20 out
of 25 instances).
\input{tabs/RFs-mx}

Although computing \iaxp's in general tends to be faster than \mxiaxp
(which is not surprising), applying our approach to computing \mxiaxp
is worthwhile when most general explanations are of concern.
This is underscored by the fact that the approach is shown to scale.
Finally, the improved bounds-based variant of the approach (both with
MaxSAT and MIP) is clearly superior to the naive propositional
encoding in terms of the overall performance.

\cref{tab:BTs-iaxp} reports the results of \mxiaxp for BTs focusing 
solely on MIP encoding. Similarly to RFs, we observe large \mxiaxp 
explanation coverage for most of the benchmarks, where 8/12 
datasets show  $12\%$ to $53\%$ coverage of the total input space. 
Additionally, we observe a net superiority of average coverage of 
\mxiaxp over \iaxp  with a factor of 2 up to 13 for 5/12 datasets.  
Overall, these empirical results validate the effectiveness  
and explanatory wide-ranging of our framework in explaining TEs.

\subsection{Additional Results: AXp \& CXp for RFwv}
\cref{tab:RFs-mx} shows the results of 
assessing the MaxSAT encoding  scalability for explaining RFww. 
	%
%The table shows results for 32 datasets.	%  
%For each dataset, we randomly pick 200 instances to be tested.
%%
%Columns {\bf m} and {\bf K} report, resp., the
%number of features and classes  in the dataset.
%%
%Columns {\bf D}, {\bf\#N} and {\bf\%A} show, resp., each tree's
%max.~depth, total number of nodes and test accuracy of an RF classifier.
%%
%Columns  {\bf \#var} and {\bf \#cl} show the average number of variables and 
%clauses of WCNF formulas encoding a forest along with any
%instance to analyze.
%%
%Column {\bf Len}  and {\bf Time} shows, resp., the average 
%length and average runtime (in seconds) for computed AXp's (resp.\ CXp's). 
%
Clearly, one can observe from the results that the MaxSAT approach for 
computing AXp's and CXp's for random forest with weighted voting (RFwm) 
demonstrates efficient performances, such that for the majority 
of datasets the runtimes  are on average less than 2 seconds per explanation, 
with a few outliers that are no longer than $\sim$28 seconds for AXp and 
$\sim$15 seconds for CXp.

%% file: tabs/RFs-iaxp.tex
%
\sisetup{parse-numbers=false,detect-all,mode=text}
\setlength{\tabcolsep}{4pt}
\let\lpr\undefined
\let\rpr\undefined
\newcommand{\lpr}{(}
\newcommand{\rpr}{)}

\begin{table*}[ht]
\centering
\resizebox{0.999\textwidth}{!}{
  \begin{tabular}{l>{\lpr}S[table-format=2,table-space-text-pre=\lpr]S[table-format=1.0,table-space-text-post=\rpr]<{\rpr}
  S[table-format=2.1]S[table-format=2.2] S[table-format=1.2]
  S[table-format=1.1]S[table-format=2.2] S[table-format=3.1] 
  S[table-format=3.2] S[table-format=2] S[table-format=3.2] S[table-format=2] S[table-format=3.2] S[table-format=2]
  S[table-format=2.3] S[table-format=4.3]}
\toprule[1.2pt]
\multirow{2}{*}{\bf Dataset} & \multicolumn{2}{c}{\multirow{2}{*}{\bf (m,~K)}} & 
\multicolumn{3}{c}{\bf  iAXp}  &
\multicolumn{3}{c}{\bf \mxiaxp} & \multicolumn{2}{c}{\bf Naive MxS} &
\multicolumn{2}{c}{\bf Bnd MxS} & \multicolumn{2}{c}{\bf Bnd MIP} &
\multicolumn{2}{c}{\bf Cov. Ratio} \\
  \cmidrule[0.8pt](lr{.75em}){4-6}
  \cmidrule[0.8pt](lr{.75em}){7-15}
  \cmidrule[0.8pt](lr{.75em}){16-17}
&  \multicolumn{2}{c}{} & {\bf Len}  & {\bf Cov\%} & {\bf Time} &
{\bf Len }  & {\bf Cov\%} & {\bf calls } &  {\bf Time$_{rc2}$} & {\bf TO$_{rc2}$} &
{\bf Time$_{rc2}$} & {\bf TO$_{rc2}$} &   {\bf Time$_{grb}$} & {\bf TO$_{grb}$} &
{\bf avg} & {\bf max} \\
\toprule[1.2pt]

ann-thyroid & 21 & 3 & 1.9 & 2.59  &  0.41 & 1.9 &  8.94  &  7.3  &  23.21  &  0  &  10.28  &  0  &  2.19  &  0  & {\bf 3.446}  &  524.711 \\
appendicitis  & 7 & 2 & 5.2 &  5.30  &  0.16 & 4.2 &  \cellcolor{midgrey!30} 18.44  &  62.7  &  282.10  &  19  &  172.11  &  11  &  97.23  &  0  & {\bf 3.478}  &  4676.519 \\
banknote  & 4 & 2 & 2.3 & \cellcolor{midgrey!30} 11.84  &  0.12 & 2.2 &  \cellcolor{midgrey!30} 24.11  &  8.3  &  428.67  &  13  &  266.35  &  1  &  20.98  &  0  & {\bf 2.037 } &  786.040 \\
breast-cancer  & 9 & 2 & 3.6 & \cellcolor{midgrey!30} 15.39  &  0.12  & 3.7 & \cellcolor{midgrey!30} 23.27  &  12.0  &  0.22  &  0  &  0.19  &  0  &  0.32  &  0  &  1.512  &  5.500 \\
bupa  & 6 & 2 & 4.7 & 0.39  &  0.15 & 4.6  &  4.20  &  18.6  &  58.62  &  2  &  38.54  &  0  &  13.02  &  0  & {\bf 10.662}  &  3128.349 \\
car  & 6 & 4 & 1.6 & \cellcolor{midgrey!30} 52.42  &  0.21 & 1.6 & \cellcolor{midgrey!30} 53.61  &  1.9  &  0.14  &  0  &  0.21  &  0  &  0.14  &  0  &  1.023  &  4.167 \\
ecoli  & 7 & 5 & 3.9 & 3.19  &  0.87 & 3.8 &  7.15  &  19.7  &  229.92  &  10  &  170.17  &  2  &  54.23  &  0  & {\bf 2.238}  &  6350.058 \\
haberman  & 3 & 2 & 1.6 & 6.54  &  0.08 & 1.7 &  8.70  &  5.0  &  0.46  &  0  &  0.41  &  0  &  0.89  &  0  &  1.330  &  11.011 \\
iris  & 4 & 3 & 2.1 & 9.65  &  0.20  & 2.1 &  9.85  &  8.1  &  0.45  &  0  &  0.65  &  0  &  0.63  &  0  &  1.021  &  2.302 \\
lupus  & 3 & 2 & 1.4 & \cellcolor{midgrey!30} 19.93  &  0.08  & 1.8 & \cellcolor{midgrey!30} 26.58  &  6.4  &  5.99  &  0  &  0.69  &  0  &  0.94  &  0  &  1.334  &  114.821 \\
new-thyroid  & 3 & 2 & 3.1 & 9.98  &  0.28 & 3.2 & \cellcolor{midgrey!30} 12.62  &  16.6  &  199.28  &  2  &  26.35  &  0  &  8.56  &  0  &  1.263  &  6048.512 \\
wine-recog  & 13 & 3 & 11.2 & 1.51  &  0.27  & 4.3 &  3.49  &  140.7  &  88.03  &  22  &  366.51  &  20  &  196.06  &  0  & {\bf 2.318}  &  364.379 \\

\bottomrule[1.2pt]
\end{tabular}
}
\caption{%
Performance evaluation of computing maximum inflated  AXp's for RFmv.
For each dataset, we randomly pick  25  instances to test.
{\bf \mxiaxp} reports the average explanation length, coverage (in \%)  and
MaxSAT oracle calls in \cref{alg:mxiaxp}.
{\bf Naive MxS}, {\bf Bnd MxS} and {\bf Bnd MIP} report
the average runtime for computing  \mxiaxp and  total timeout
tests, resp., for naive MaxSAT, Bounds-based MaxSAT and MIP
encodings.
Column {\bf Cov. Ratio} reports, resp., the average and maximum ratio
between domain coverage of {\bf \mxiaxp} and  {\bf \iaxp}.
Coverage ratios higher than a factor of 2 are highlighted in bold text and
\mxiaxp coverages greater than $10\%$ are highlighted in grey.
}
\label{tab:RFs-iaxp}
\end{table*}

%% file: tabs/BTs-iaxp.tex
\sisetup{parse-numbers=false,detect-all,mode=text}
\setlength{\tabcolsep}{3pt}
\let\lpr\undefined
\let\rpr\undefined
\newcommand{\lpr}{(}
\newcommand{\rpr}{)}

\begin{table}[ht]
\centering
\resizebox{1.0\columnwidth}{!}{
  \begin{tabular}{l%>{\lpr}S[table-format=2.0,table-space-text-pre=\lpr]S[table-format=3.0,table-space-text-post=\rpr]<{\rpr}
  S[table-format=2.2] S[table-format=1.2]
  S[table-format=2.2] S[table-format=3.1] S[table-format=2.2]
  S[table-format=2.2] S[table-format=5.1]}
\toprule[1.2pt]
\multirow{2}{*}{\bf Dataset} & \multicolumn{2}{c}{\bf  iAXp} & \multicolumn{3}{c}{\bf \mxiaxp} & \multicolumn{2}{c}{\bf Cov. Ratio} \\
  \cmidrule[0.8pt](lr{.75em}){2-3}
  \cmidrule[0.8pt](lr{.75em}){4-6}
  \cmidrule[0.8pt](lr{.75em}){7-8}
& {\bf Cov\%} & {\bf Time} &  {\bf Cov\%} & {\bf calls } & {\bf Time$_{grb}$} & {\bf avg} & {\bf max} \\
\midrule[1.2pt]

ann-thyroid  &  1.34  &  0.27  &  \cellcolor{midgrey!30} 17.77  &  2.0  &  1.56  & {\bf \num{13.27}}  &  59.4 \\
appendicitis  &  6.59  &  0.09  &  \cellcolor{midgrey!30} 24.26  &  10.4  &  0.90  &  {\bf \num{3.68}}  &  587.7 \\
banknote  & \cellcolor{midgrey!30} 18.21  &  1.13  & \cellcolor{midgrey!30} 21.33  &  5.5  &  15.86  &  1.17  &  22.3 \\
breast-cancer  &  0.38  &  1.14  &  2.21  &  17.9  &  10.61  &  {\bf \num{5.81}}  &  131.6 \\
bupa  &  0.49  &  1.26  &  2.78  &  20.6  &  18.92  &  {\bf 5.63}  &  98379.4 \\
car  & \cellcolor{midgrey!30} 26.54  &  0.14  &  \cellcolor{midgrey!30} 26.91  &  2.2  &  0.18  &  1.01  &  22.5 \\
ecoli  &  2.92  &  6.88  &  5.29  &  17.8  &  83.84  &  1.81  &  743.8 \\
haberman  &  1.40  &  0.56  &  2.71  &  6.0  &  3.30  &  1.93  &  36.2 \\
iris  & \cellcolor{midgrey!30} 20.63  &  0.13  &  \cellcolor{midgrey!30} 22.29  &  3.2  &  0.38  &  1.08  &  22.5 \\
lupus  & \cellcolor{midgrey!30} 16.56  &  0.07  &  \cellcolor{midgrey!30} 19.47  &  3.6  &  0.28  &  1.18  &  7.7 \\
new-thyroid  &  4.63  &  0.41  &  7.39  &  9.7  &  2.73  &  1.60  &  246.7 \\
%rice  & \cellcolor{midgrey!30} 14.12  &  0.36  &  \cellcolor{midgrey!30} 20.10  &  28.8  &  13.20  &  1.42  &  10214.0 \\
%shuttle  &  7.60  &  2.91  &  8.14  &  10.2  &  18.66  &  1.07  &  168616.1 \\
wine-recog  &  6.18  &  0.26  &  \cellcolor{midgrey!30} 14.02  &  16.1  & 3.80  &  {\bf \num{2.27}}  &  212.8 \\

\bottomrule[1.2pt]
\end{tabular} }
\caption{%
Performance evaluation of computing \mxiaxp's of BTs.}
\label{tab:BTs-iaxp}
\end{table}

%% file: tabs/RFs-mx.tex
\setlength{\tabcolsep}{5pt}
\let\lpr\undefined
\let\rpr\undefined
\newcommand{\lpr}{(}
\newcommand{\rpr}{)}

\begin{table*}[ht]
\centering
\resizebox{0.75\textwidth}{!}{
  \begin{tabular}{l>{\lpr}S[table-format=2,table-space-text-pre=\lpr]S[table-format=2,table-space-text-post=\rpr]<{\rpr} 
  S[table-format=1]S[table-format=4.0]S[table-format=3]
  >{\lpr}S[table-format=5.0,table-space-text-pre=\lpr]S[table-format=5.0,table-space-text-post=\rpr]<{\rpr} 
  S[table-format=2.1]S[table-format=2.2]
  S[table-format=2.1]S[table-format=2.2]}
\toprule[1.2pt]
\multirow{2}{*}{\bf Dataset} & \multicolumn{2}{c}{\multirow{2}{*}{\bf (m,~K)}}  & \multicolumn{3}{c}{\bf RF} & \multicolumn{2}{c}{\bf MxS Enc}  & %\multicolumn{4}{c}{\bf MaxSAT oracle} & 
\multicolumn{2}{c}{\bf AXp} & \multicolumn{2}{c}{\bf CXp}  \\
  \cmidrule[0.8pt](lr{.75em}){4-6}
  \cmidrule[0.8pt](lr{.75em}){7-8}
  \cmidrule[0.8pt](lr{.75em}){9-10}
  \cmidrule[0.8pt](lr{.75em}){11-12}
  %%%%%
  %
& \multicolumn{2}{c}{} & {\bf D}  & {\bf \#N} & {\bf \%A} & {\bf \#var} & {\bf \#cl} &  {\bf Len} & {\bf Time} &  {\bf Len} & {\bf Time} \\
\toprule[1.2pt]

ann-thyroid  &  21  &  3  &  4  &  1094  &  97  &  3373  &  7582  &  1.9  &  1.53  &  1.8  &  0.71 \\
appendicitis  &  7  &  2  &  6  &  980  &  90  &  3037  &  7435  &  4.1  &  0.24  &  2.7  &  0.27 \\
banknote  &  4  &  2  &  5  &  1348  &  98  &  4419  &  11041  &  2.2  &  0.21  &  1.4  &  0.16 \\
%biodegradation  &  41  &  2  &  4  &  1332  &  88  &  4455  &  10416  &  15.9  &  5.19  &  4.7  &  3.18 \\
ecoli  &  7  &  5  &  5  &  1962  &  87  &  6259  &  14332  &  3.4  &  0.94  &  1.1  &  0.47 \\
glass2  &  9  &  2  &  6  &  1494  &  87  &  4713  &  12026  &  4.6  &  0.52  &  1.9  &  0.34 \\
heart-c  &  13  &  2  &  5  &  1964  &  83  &  5726  &  13774  &  6.0  &  1.11  &  2.0  &  0.48 \\
%ionosphere  &  34  &  2  &  5  &  1056  &  84  &  3566  &  8763  &  19.6  &  0.95  &  6.6  &  0.71 \\
iris  &  4  &  3  &  6  &  734  &  93  &  2047  &  4618  &  2.2  &  0.09  &  1.3  &  0.07 \\
karhunen  &  64  &  10  &  4  &  1546  &  91  &  7475  &  13600  &  25.6  &  27.61  &  8.1  &  15.21 \\
magic  &  10  &  2  &  6  &  4894  &  85  &  17624  &  47487  &  6.0  &  12.67  &  1.8  &  7.35 \\
mofn-3-7-10  &  10  &  2  &  6  &  4384  &  92  &  11416  &  28037  &  3.2  &  1.54  &  1.9  &  1.06 \\
new-thyroid  &  5  &  3  &  5  &  908  &  100  &  2775  &  6418  &  2.9  &  0.19  &  1.2  &  0.14 \\
pendigits  &  16  &  10  &  6  &  6014  &  95  &  23303  &  49062  &  9.7  &  28.33  &  1.6  &  7.09 \\
phoneme  &  5  &  2  &  6  &  4566  &  84  &  15730  &  42173  &  3.0  &  4.00  &  1.6  &  2.50 \\
ring  &  20  &  2  &  5  &  1980  &  90  &  7051  &  18004  &  10.5  &  4.22  &  2.6  &  1.55 \\
segmentation  &  19  &  7  &  4  &  984  &  88  &  3444  &  7759  &  8.0  &  1.31  &  3.3  &  0.59 \\
shuttle  &  9  &  7  &  3  &  736  &  99  &  2155  &  4484  &  2.3  &  0.74  &  2.4  &  0.47 \\
sonar  &  60  &  2  &  5  &  1320  &  88  &  4682  &  11711  &  32.1  &  2.96  &  8.0  &  1.22 \\
spambase  &  57  &  2  &  4  &  1370  &  92  &  4747  &  11148  &  14.8  &  5.49  &  4.9  &  2.51 \\
spectf  &  44  &  2  &  5  &  1180  &  87  &  3867  &  9393  &  18.9  &  1.92  &  6.6  &  1.22 \\
texture  &  40  &  11  &  5  &  2844  &  89  &  12488  &  25973  &  23.3  &  20.99  &  3.8  &  6.03 \\
threeOf9  &  9  &  2  &  3  &  434  &  100  &  970  &  1956  &  1.0  &  0.07  &  1.0  &  0.05 \\
twonorm  &  20  &  2  &  3  &  750  &  92  &  2579  &  6059  &  9.9  &  0.82  &  3.6  &  0.54 \\
vowel  &  13  &  11  &  6  &  5096  &  90  &  20392  &  48230  &  7.8  &  15.70  &  2.8  &  8.87 \\
waveform-21  &  21  &  3  &  4  &  1550  &  82  &  5667  &  12784  &  8.3  &  5.51  &  2.7  &  2.94 \\
waveform-40  &  40  &  3  &  4  &  1550  &  81  &  5766  &  13057  &  10.0  &  9.47  &  3.3  &  4.34 \\
wdbc  &  30  &  2  &  4  &  1028  &  95  &  3560  &  8637  &  12.3  &  1.27  &  5.0  &  0.57 \\
wine-recog  &  13  &  3  &  3  &  592  &  97  &  1976  &  4369  &  4.4  &  0.29  &  1.9  &  0.19 \\
wpbc  &  33  &  2  &  5  &  1234  &  76  &  4290  &  10844  &  20.4  &  2.74  &  7.9  &  1.33 \\
xd6  &  9  &  2  &  6  &  4170  &  100  &  10731  &  26541  &  3.0  &  0.80  &  1.5  &  0.53 \\

\bottomrule[1.2pt]
\end{tabular}
}
\caption{%
	Assessing the  MaxSAT encoding  scalability for explaining RFwv. 
	The table shows results for 32 datasets.	%  
	For each dataset, we randomly pick 200 instances to be tested.
	Columns {\bf m} and {\bf K} report, resp., the
	number of features and classes  in the dataset.
	Columns {\bf D}, {\bf\#N} and {\bf\%A} show, resp., each tree's
	max.~depth, total number of nodes and test accuracy of an RF classifier.
	Columns  {\bf \#var} and {\bf \#cl} show the average number of variables and 
	clauses of WCNF formulas encoding a forest along with any
	instance to analyze.
	Column {\bf Len}  and {\bf Time} shows, resp., the average 
	length and average runtime (in seconds) for computed AXp's (resp.\ CXp's). 
}
\label{tab:RFs-mx}
\end{table*}

%% file: conc.tex
\section{Conclusions} \label{sec:conc}
This paper takes a step further in the quest of finding most succinct 
and general explanations for (complex) ML models. This is achieved 
by formalizing 
the concept of {\it maximum inflated abductive explanation} (\mxiaxp),
which can be considered as the most general abductive explanations.
Furthermore, the paper proposes an elegant approach for the rigorous
computation of maximum inflated 
explanations, and demonstrates the broader coverage of such 
explanations in comparison with maximal (subset) inflated abductive  
explanations. %~\cite{iisms-aaai24}.
The experimental results validate the practical interest of computing
\emph{maximum} \iaxp explanations.

%% file: ack.tex
\section{Acknowledgments}
This work was supported in part by the National Research Foundation, 
Prime Minister’s Office, Singapore under its Campus for Research 
Excellence and Technological Enterprise (CREATE) programme,
by the Spanish Government under grant PID2023-152814OB-I00 and  
ICREA starting funds,
and by the Australian Research Council through the OPTIMA ITTC IC200100009.

Finally, we thank Rayman Tang for the discussions at an early stage 
of this work.

%% file: replbib.tex
% RequiredL: \usepackage{etoolbox}
%\providetoggle{mkbbl}
\newtoggle{mkbbl}
% Contents if using bibtex: "\settoggle{mkbbl}{true}"
% Contents if inputing pre-generated file: "\settoggle{mkbbl}{false}"

%% file: togbbl.tex
\settoggle{mkbbl}{false}